\documentclass[sigconf]{acmart}

\usepackage{graphicx}

\usepackage{times}
\usepackage{latexsym}
\usepackage{amsmath}
\usepackage[T1]{fontenc}
\usepackage[utf8]{inputenc}
\usepackage{microtype}

\usepackage{graphicx}
\usepackage{natbib}
\usepackage{helvet}  
\usepackage{courier}  
\usepackage{multirow}
\usepackage{xcolor}
\usepackage{booktabs}
\usepackage[export]{adjustbox}
\usepackage{xcolor,colortbl}
\definecolor{bubbles}{rgb}{0.91, 1.0, 1.0}

\definecolor{Gray}{gray}{0.85}
\definecolor{LightCyan}{rgb}{0.88,1,1}
\definecolor{lightthulianpink}{rgb}{0.9, 0.56, 0.67}
\definecolor{mypink3}{cmyk}{0, 0.7808, 0.4429, 0.1412}
\newcolumntype{a}{>{\columncolor{Gray}}c}
\newcolumntype{b}{>{\columncolor{white}}c}
\definecolor{blizzardblue}{rgb}{0.67, 0.9, 0.93}

\usepackage{amsmath,amsfonts,bm}
\usepackage{xcolor}
\usepackage{graphicx}
\usepackage{bm}

\theoremstyle{definition}
\newtheorem{definition}{Definition}[section]
\usepackage{amsmath}
\usepackage{balance}
\usepackage{enumitem}

\usepackage{amsmath}
\usepackage{mathtools}
\usepackage{amsthm}

\usepackage{array}
\usepackage{xspace}

\AtBeginDocument{%
  \providecommand\BibTeX{{%
    \normalfont B\kern-0.5em{\scshape i\kern-0.25em b}\kern-0.8em\TeX}}}

\setcopyright{acmlicensed}
\copyrightyear{2024}
\acmYear{2024}
\setcopyright{acmlicensed}\acmConference[CCS '24]{Proceedings of the 2024
ACM SIGSAC Conference on Computer and Communications Security}{October
14--18, 2024}{Salt Lake City, UT, USA}
\acmBooktitle{Proceedings of the 2024 ACM SIGSAC Conference on Computer
and Communications Security (CCS '24), October 14--18, 2024, Salt Lake City,
UT, USA}
\acmDOI{10.1145/3658644.3690217}
\acmISBN{979-8-4007-0636-3/24/10}

\begin{document}

\newcommand{\icon}{\raisebox{8pt}{\includegraphics[valign=c, width=1.2em]{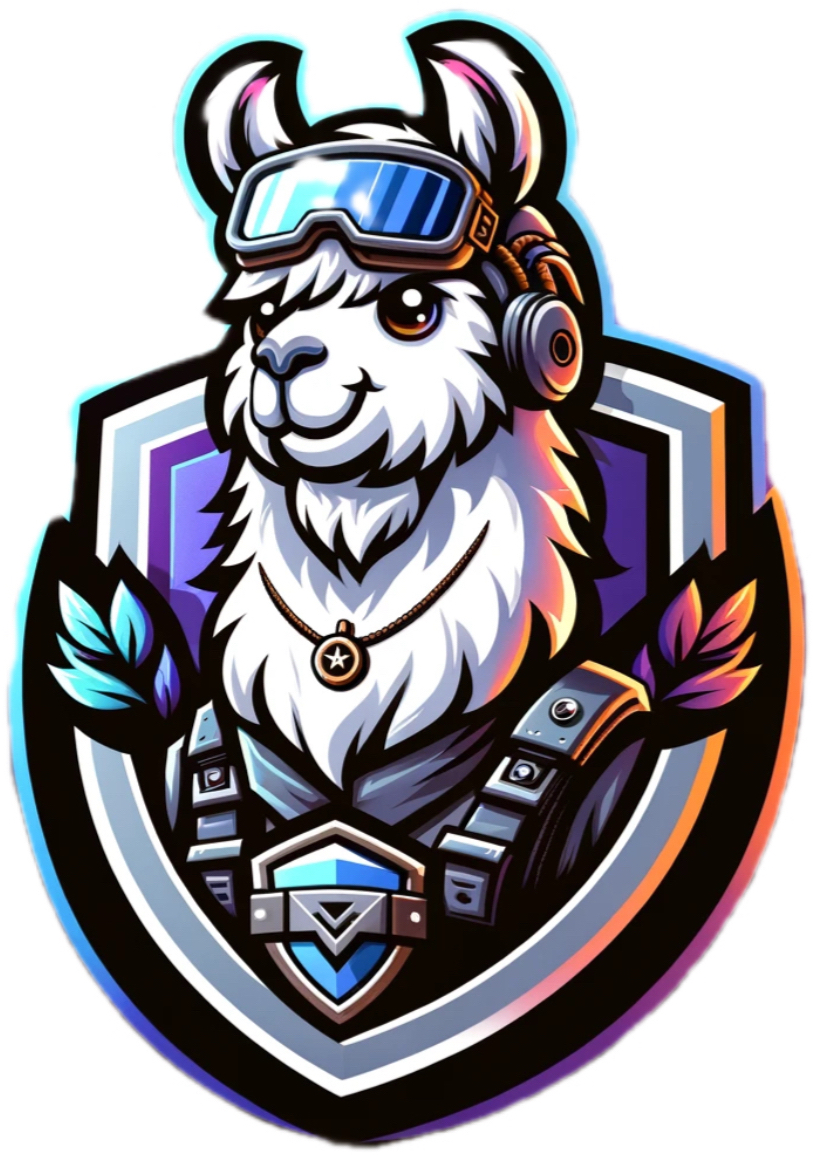}}\xspace}

\title{\icon A Causal Explainable Guardrails for Large Language Models}

\author{Zhixuan Chu}
\authornote{These authors contributed equally to this work.}
\authornote{Corresponding author. The author is with the State Key Laboratory of Blockchain and Data Security \& Hangzhou High-Tech Zone (Binjiang) Institute of Blockchain and Data Security, Hangzhou, China.}
\orcid{}
\affiliation{%
  \institution{Zhejiang University}
  \streetaddress{}
  \city{Hangzhou}
  \state{}
  \country{China}
  \postcode{}
}
\email{zhixuanchu@zju.edu.cn}

\author{Yan Wang}
\authornotemark[1]
\affiliation{%
  \institution{Ant Group}
  \streetaddress{}
  \city{Hangzhou}
  \state{}
  \country{China}
  \postcode{}
}
\email{luli.wy@antgroup.com}

\author{Longfei Li}
\affiliation{%
  \institution{Ant Group}
  \streetaddress{}
  \city{Hangzhou}
  \state{}
  \country{China}
  \postcode{}
}
\email{longyao.llf@antgroup.com}

\author{Zhibo Wang}
\affiliation{%
  \institution{Zhejiang University}
  \streetaddress{}
  \city{Hangzhou}
  \state{}
  \country{China}
  \postcode{}
}
\email{zhibowang@zju.edu.cn}

\author{Zhan Qin}
\affiliation{%
  \institution{Zhejiang University}
  \streetaddress{}
  \city{Hangzhou}
  \state{}
  \country{China}
  \postcode{}
}
\email{qinzhan@zju.edu.cn}

\author{Kui Ren}
\affiliation{%
  \institution{Zhejiang University}
  \streetaddress{}
  \city{Hangzhou}
  \state{}
  \country{China}
  \postcode{}
}
\email{kuiren@zju.edu.cn}

\renewcommand{\shortauthors}{Zhixuan Chu and Yan Wang, et al.}

\begin{abstract}
Large Language Models (LLMs) have shown impressive performance in natural language tasks, but their outputs can exhibit undesirable attributes or biases. Existing methods for steering LLMs toward desired attributes often assume unbiased representations and rely solely on steering prompts. However, the representations learned from pre-training can introduce semantic biases that influence the steering process, leading to suboptimal results. We propose LLMGuardrail, a novel framework that incorporates causal analysis and adversarial learning to obtain unbiased steering representations in LLMs. LLMGuardrail systematically identifies and blocks the confounding effects of biases, enabling the extraction of unbiased steering representations. Additionally, it includes an explainable component that provides insights into the alignment between the generated output and the desired direction. Experiments demonstrate LLMGuardrail's effectiveness in steering LLMs toward desired attributes while mitigating biases. Our work contributes to the development of safe and reliable LLMs that align with desired attributes. 
\end{abstract}




\maketitle

\section{Introduction}

Large Language Models (LLMs) have demonstrated remarkable capabilities in natural language understanding and generation, enabling a wide range of applications such as dialogue systems, clinical report generation, professional agents, recommendation systems, and so on \cite{liu2024survey,guan2024intelligent,chu2024professional,chu2024llm,wang2024llmrg}. However, the training of these models on massive web-scraped corpora has led to the manifestation of undesirable behaviors, such as generating offensive, toxic, or false outputs \cite{huang2023survey,wen2023unveiling,chu2024sora}. Addressing these issues is crucial, as content safety, fairness, toxicity, harmfulness, and factuality demand rigorous consideration, especially in an era where language models are increasingly deployed in high-stakes applications and user-facing systems.

To address these challenges, recent research efforts have focused on developing methods to steer the output of LLMs toward desired attributes or concepts. Several approaches have been proposed to control and steer the generation process, aiming to mitigate these harmful behaviors and improve the overall quality and safety of generated content. Fine-tuning on carefully curated datasets has been a common technique to enhance content safety and regulate outputs. By training language models on datasets that have been filtered and cleaned to remove offensive, biased, or misleading content, researchers aim to reduce the likelihood of the model generating such harmful outputs.
In addition to fine-tuning, other techniques have been explored to further improve the generation process. Reinforcement learning from human feedback (RLHF) \cite{ouyang2022training} is one such approach, where the language model is trained using feedback from human annotators who evaluate the quality and safety of generated outputs. Another promising approach is reinforcement learning from AI feedback (RLAIF) \cite{lee2023rlaif}, which extends the concept of RLHF by using an AI system to provide feedback instead of human annotators. However, they all require huge annotation and computation resources. Furthermore, the training process often involves a human or AI annotator providing feedback and guidance to the language model. This raises the possibility that some form of deception or manipulation could be introduced during the training process, either intentionally or unintentionally \cite{tarsney2024deception,tan2024large,qi2023fine}. In addition, these methods often lack explainability and interpretability, leading to inconsistent performance and limited generalizability \cite{liao2023ai,zhao2024explainability,hickling2023explainability}. 

\begin{figure}[t]
    \centering
    \includegraphics[width=0.6\columnwidth]{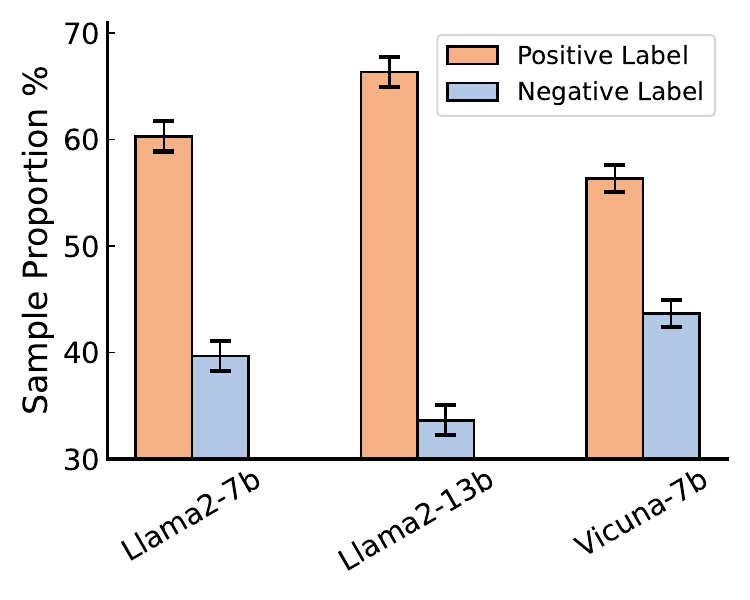}
    \vspace{-4mm}
    \caption{Proportion of representations of semantic prompts that implicitly encode positive or negative directions for semantically neutral prompts without explicit steering prompt, across different language models. The varying proportions of positive and negative directions learned by the probing classifier, even in the absence of steering prompts, demonstrate the presence of inherent biases in the representations of semantic prompts due to differences in pre-training data. This observation supports the existence of a direct edge from the semantic direction representation $R^{cd}$ of the semantic prompt to the direction representation $R_{+}/R_{-}$, as discussed in the causal analysis section.}
    \vspace{-4mm}
    \label{fig:model_bias}
\end{figure}

Recent studies have shed light on the rich semantic information encoded within LLM representations, including specific information \cite{gurnee2023language}, concepts \cite{mikolov2013linguistic,zou2023representation}, or attributes \cite{bolukbasi2016man,li2024inference}. Building on these findings, activation engineering has emerged, which induces desired outputs in frozen LLMs by injecting crafted activation vectors during forward passes. This approach holds promise for controlling the generation process in a more interpretable and fine-grained manner. Existing approaches for steering LLMs typically involve constructing positive and negative prompts that represent the desired and undesired attributes, respectively. By comparing the activations of the model for these prompts, a steering vector is obtained, which is then used to guide the model's output during inference. While these methods have shown promising results, they often rely on the assumption that the constructed prompts are sufficient to capture the desired steering direction and that the model's representations are unbiased. However, a closer examination reveals that the representations learned by LLMs during pre-training can introduce biases that influence the steering process. As shown in Figure \ref{fig:model_bias}, the semantic context of the prompts used for steering can implicitly encode biases, even without explicit steering prompts. Consequently, the obtained steering vectors may be confounded by these inherent biases, leading to suboptimal or unintended steering results. This observation highlights a critical gap in current steering approaches: the need to account for and mitigate the influence of semantic biases on the steering process. Without addressing this issue, steering methods may inadvertently perpetuate or even amplify existing biases in the model's outputs, undermining efforts to improve the safety and reliability of LLMs. Therefore, simply structuring a controlled trial superficially is not enough to get an unbiased steering representation.

To address these limitations, we propose a novel framework called LLMGuardrail that incorporates causal analysis to obtain unbiased steering representations for LLMs. Our framework aims to disentangle the influence of biases from the steering process by employing debiasing techniques. By systematically identifying and blocking the confounding effects of biases, LLMGuardrail enables the extraction of steering representations that accurately capture the desired attributes. The key contributions of this work are as follows:

\begin{itemize}[leftmargin=*]
 \item A causal analysis of the steering process in LLMs, identifying the confounding effects of semantic prompt and their impact on steering representations. We provide a theoretical formulation of the problem and discuss the limitations of existing methods.
 \item  A novel framework for obtaining unbiased steering representations in LLMs. Our framework employs adversarial learning techniques to disentangle the influence of semantic biases from the steering process.
 \item An explainable component that provides insights into the alignment between the generated output and the desired direction, enhancing the interpretability of the steering process.
 \item  Comprehensive experiments and analysis demonstrating the effectiveness of LLMGuardrail in steering LLMs toward desired attributes while mitigating the influence of biases. We evaluate our framework on various benchmark datasets and compare its performance with existing methods.
\end{itemize}

\begin{figure*}[t]
    \centering
    \includegraphics[width=2\columnwidth]{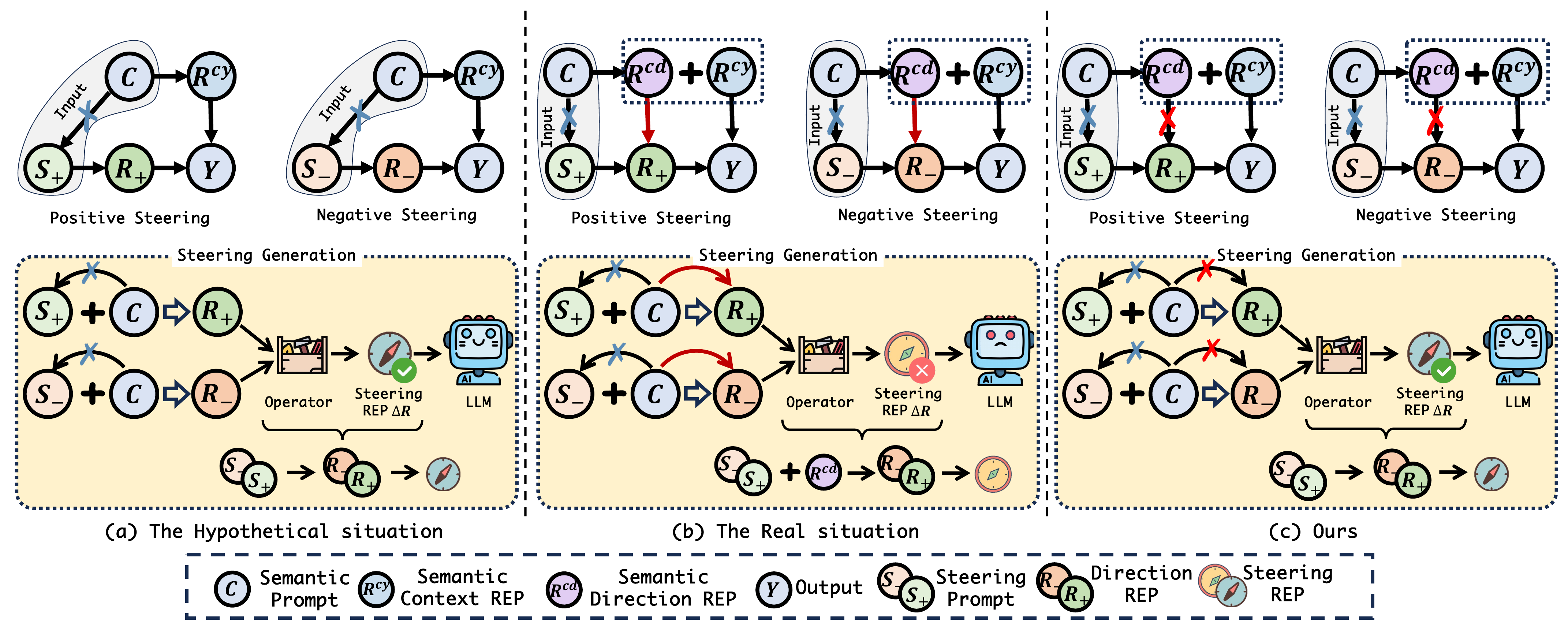}
    \vspace{-4mm}
    \caption{The causal analysis of our proposed LLMGuardrail. (a) The Hypothetical situation: The constructed pair of prompts can block the edge from the semantic prompt $C$ to the steering prompt $S$ and the direction representation $R^{+}/R^{-}$. (b) The Real situation: In addition to the edge from the semantic prompt $C$ to the steering prompt $S$, there is a direct edge from the semantic direction representation $R^{cd}$ of the semantic prompt to the direction representation $R^{+}/R^{-}$. (c) Ours: We need to block the edge from the semantic direction representation $R^{cd}$ of the semantic prompt to the direction representation $R^{+}/R^{-}$. The direction representation $R^{+}/R^{-}$ is only influenced by the steering prompt $S^{+}/S^{-}$, enabling us to obtain an unbiased steering representation $\Delta R$ through steering engineering.}

    \label{fig:causal_graph}
\end{figure*}

\section{Background}

\subsection{Representations in Large Language Models}
 
Large language models (LLMs) have demonstrated remarkable capabilities in natural language processing tasks, generating human-like text and exhibiting a broad understanding of language. Central to their success is the rich set of representations they learn during training, which encode various concepts, attributes, and semantic information. Extensive research has been dedicated to understanding the representations learned by large language models (LLMs) and how they encode various concepts and attributes \cite{zou2023representation}. LLM representations refer to the patterns of activations across the model's parameters that correspond to specific semantic concepts, properties, or features. These representations act as the model's internal codification of the information learned from the training data. Recent studies have provided compelling evidence that LLM representations contain rich semantic information, including abstract concepts like space and time \cite{gurnee2023language}, as well as more granular attributes related to truthfulness, toxicity, bias, and harmfulness of the generated text \cite{chen2024inside,li2024inference,azaria2023internal,burns2022discovering}. In addition, linear classifier probes, which are trained to predict input properties from intermediate layers of a network, have successfully identified representations of concepts \cite{alain2016understanding,belinkov2022probing}. Latent space analysis has enabled researchers to locate or edit factual associations within LLMs \cite{hernandez2023inspecting,meng2022locating,zhong2023mquake}. The ability to locate and manipulate these representations opens up avenues for controlling and steering the model's outputs, enabling the development of principled methods for mitigating harmful behaviors and enhancing the interpretability and trustworthiness of large language models.

\subsection{LLM Activation Engineering}

Activation engineering is a set of techniques that modify the internal activations of a pre-trained language model during inference to steer its output in a desired direction. The main idea is to identify and manipulate specific activations or attention heads associated with particular attributes or behaviors to control the model's generation process. Early approaches, like the Plug-and-Play Language Model \cite{dathathri2019plug}, use a separate classifier to detect target attributes in the generated text and perturb the language model's activations accordingly, encouraging it to generate text that aligns with the desired attribute. Recent work focuses on extracting latent steering vectors from a frozen language model, which can be added to the activations during inference to steer the model's completions toward specific goals, such as achieving high BLEU scores \cite{subramani2022extracting} or generating truthful statements \cite{li2024inference}. Instead of requiring additional optimization or labeled data, Activation Addition \cite{turner2023activation} takes activation differences resulting from pairs of prompts. To avoid the identification of an opposite behavior, \cite{jorgensen2023improving} takes the average of activations associated with a target dataset and then subtracts the mean of all training activations, resulting in effective steering vectors. The growing interest in activation engineering stems from the desire to control and improve the performance of large language models on specific tasks or domains by identifying and manipulating the relevant activations.

\subsection{Causal Inference}

\begin{figure}[t]
    \centering
    \includegraphics[width=0.65\columnwidth]{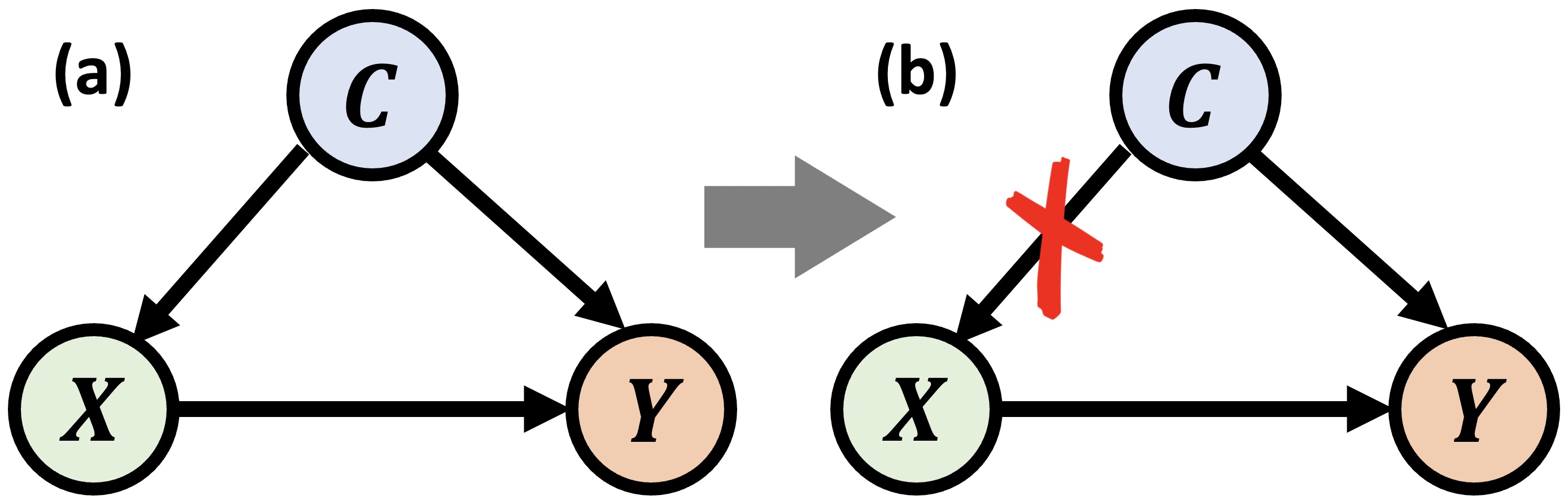}
    \vspace{-2mm}
    \caption{The causal graph and backdoor adjustment.}
    \vspace{-6mm}
    \label{fig:basic_causal_graph}
\end{figure}

Causal inference \cite{pearl2000causality,rubin2005causal,yao2021survey,chu2023causal,cui2020causal} has been an attractive research topic for a long time as it provides an effective way to uncover causal relationships in real-world problems. Nowadays, causal inference has shown potential in enhancing LLMs from a causal view in improving the LLMs' reasoning capacity \cite{chu2023data,gao2023chatgpt,zhang2022rock,chen2024learning,liu2022incorporating}, addressing fairness issues in LLMs \cite{stanovsky2019evaluating,ding2022word,meade2021empirical}, and safety \cite{bao2021defending,cao2022can,liu2023hallusionbench}, complementing LLMs with explanations \cite{liu2023aligning,zhao2024explainability,belrose2023eliciting,wan2024bridging}, and handling multimodality \cite{ko2023large,pawlowski2023answering}. In causal inference, if a variable is the common cause of two variables, it is called the \textit{confounder}. The confounder will induce a spurious correlation between these two variables to disturb the recognition of the causal effect between them. As shown in Figure \ref{fig:basic_causal_graph}(a), $X\rightarrow Y$ denotes that $X$ is the cause of $Y$. $C$ is the cause of both $X$ and $Y$. Thus, it is a confounder that will induce a spurious correlation between $X$ and $Y$ to disturb the recognition of the causal effect between them. In particular, such spurious correlation is brought by the backdoor path created by the confounder. Formally, a backdoor path between $X$ and $Y$ is defined as any path from $X$ to $Y$ that starts with an arrow pointing into $X$. For example, the path $X \leftarrow C \rightarrow Y$ is a backdoor path. If we want to deconfound two variables $X$ and $Y$ to calculate the true causal effect, we should block every backdoor path between them~\cite{PearlMackenzie18}. For example, in Figure~\ref{fig:basic_causal_graph}(b), we should block $X \leftarrow C \rightarrow Y$ to get the causal effect between $X$ and $Y$.

\section{Causal Analysis}

\subsection{The Hypothetical Situation}

Recent activation engineering-based work \cite{burns2022discovering,zou2023representation,turner2023activation,jorgensen2023improving,li2024inference} utilize the steering vectors to control the direction of LLM output by constructing the randomized controlled trials. For example, some work \cite{burns2022discovering,zou2023representation,turner2023activation} construct a pair of natural-language prompts ($p_+$, $p_-$), where $p_+$ represents the attribute we wish output text to emphasize and $p_-$ represents its opposite. $R_+/R_-$ is the representation for the prompt $p_+/p_-$. The difference $\Delta R$ is a new steering vector that (intuitively) captures the difference between a prompt with the attribute and without it. To obtain a steering vector, they perform a forward pass on each prompt, record the activations at the given location in each pass, take the difference, and then finally rescale this difference in activations by an ``injection coefficient'' $\beta$. To steer, they add the resulting steering representation to the original representations and allow the forward pass to continue, and obtain the steered output.

To systemically analyze this series of methods, we give the formal definitions. Now, let's further break down and analysis of the constructed prompt pairs. In fact, the input prompt is comprised of the semantic prompt $C$ and the steering prompt $S$. For example, in the \cite{turner2023activation}, ``I love talking
about weddings'' is the positive prompt and ``I hate talking about weddings'' is the negative prompt. In these two examples,  ``love'' and ``hate'' are the steering prompts, which control the directions of output. The ``I ... talking about weddings'' is the same for this pair of prompts, which only contains the semantic information. Therefore, they assume this is a randomized controlled trial, where there are two groups, i.e., treatment group ``love'' and control group ``hate''. As shown in Figure \ref{fig:causal_graph} (a), the semantic prompt $C$ is a confounder, that can affect the steering prompt $S$ and output $Y$. In this hypothetical situation, they construct the randomized controlled trial by creating the pairs, i.e., positive steering prompt plus the same semantic prompt and negative steering prompt plus the same semantic prompt. The only difference is the steering prompt since the semantic prompts are the totally same. Based on this assumption, the edge from the semantic prompt to the steering prompt is blocked. Therefore, they can get the difference in activations, which can intervene in the generated output. The strength of this steering vector is called the treatment effect in causal inference. On the face of it, this is a perfect randomized controlled trial, where only the steering prompt can affect the direction representation $R_+/R_-$. The combination of direction representation $R_+/R_-$ and the semantic prompt together have an influence on the generated output $Y$. In this case, the steering vector can be obtained by the operation between positive direction representation $R_+$ and negative direction representation $R_-$.

\subsection{The Real Situation}

Before delving into the analysis of the real situation, we first provide formal definitions for the various types of representations involved in the steering process.

\begin{definition}[Direction Representation $R_{+/-}$]
A direction representation $R_{+/-}$ is a representation that solely affects the direction of the output with respect to specific attributes such as truthfulness, bias, harmfulness, or toxicity. It is learned from the steering prompt and should be independent of the semantic context of the output.
\end{definition}

\begin{definition}[Semantic Context Representation $R^{cy}$]
A semantic context representation $R^{cy}$ is a representation learned from the semantic prompt, which contains information about the context of the output. It does not provide guidance for the direction of the output with respect to specific attributes.
\end{definition}

\begin{definition}[Semantic Direction Representation $R^{cd}$]
A semantic direction representation $R^{cd}$ is a representation learned from the semantic prompt that implicitly influences the direction of the output with respect to specific attributes. This influence stems from associations learned during the pre-training of the large language model, which may introduce biases related to the desired attributes.
\end{definition}

\begin{definition}[Steering Representation $\Delta R$]
A steering representation $\Delta R$ is a representation that stands for the direction of the output with respect to specific attributes such as truthfulness, bias, harmfulness, or toxicity. It is obtained by computing the difference between the positive direction representation $R_{+}$ and the negative direction representation $R_{-}$, which are learned from the corresponding steering prompts. The steering representation should be independent of the semantic direction representation $R^{cd}$ to ensure unbiased steering of the output.
\end{definition}

Based on the above definitions, the assumed randomized controlled trial in the hypothetical situation is not qualified because it fails to consider the influence of confounders, specifically the semantic prompt, on the direction representation $R_{+}/R_{-}$ due to biases in the pre-training of the large language model (LLM). As a result, the obtained steering vector is biased. As proven in the introduction section, the semantic prompt $C$ affects not only the content of the output $Y$ but also its direction. We can further break down the semantic prompt into two parts, i.e., the semantic context representation $R^{cy}$, which influences the content of the output $Y$, and the semantic direction representation $R^{cd}$, which influences the direction representation $R_{+}/R_{-}$.

In addition to the edge from the semantic prompt $C$ to the steering prompt $S$ and the direction representation $R_{+}/R_{-}$, there is a direct edge from the semantic direction representation $R^{cd}$ of the semantic prompt to the direction representation $R_{+}/R_{-}$. The constructed pair of prompts can only block the edge from the semantic prompt $C$ to the steering prompt $S$ and the direction representation $R_{+}/R_{-}$, but it ignores the edge from the semantic direction representation $R^{cd}$ of the semantic prompt to the direction representation $R_{+}/R_{-}$. As shown in Figure \ref{fig:causal_graph} (b), both the steering prompt $S_{+}/S_{-}$ and the semantic direction representation $R^{cd}$ affect the direction representation $R_{+}/R_{-}$. Consequently, the obtained steering vector $\Delta R$ is biased by the semantic direction representation $R^{cd}$ of the semantic prompt, which is not solely affected by the constructed steering prompt $S$.

This bias originates from the biases in the pre-training data. For example, consider the semantic prompt ``I ... talking about weddings''. Due to biases present in the pre-training data, the LLM may have learned to associate weddings with positive sentiments such as love, happiness, and celebration. As a result, the semantic direction representation $R^{cd}$ learned from this prompt may implicitly suggest a positive direction (e.g., ``love'') for the output, even if the explicit steering prompt is not provided. As shown in Figure \ref{fig:model_bias}, this edge can be visually observed experimentally. The implicit influence of the semantic direction representation $R^{cd}$ on the output direction can be problematic when attempting to steer the LLM's output toward a desired attribute. If the steering representation $\Delta R$ is not independent of $R^{cd}$, the resulting output may be biased by the inherent associations learned during pre-training, leading to suboptimal or unintended results. To ensure unbiased steering of the output, it is crucial to disentangle the steering representation $\Delta R$ from the semantic direction representation $R^{cd}$.

\subsection{Causal Analysis of Our Solutions}
Based on the causal analysis presented in the previous sections, we propose a solution called LLMGuardrail to address the bias introduced by the semantic direction representation $R^{cd}$ in the steering process. As shown in Figure \ref{fig:causal_graph} (c), in addition to constructing pairs of prompts (positive and negative steering prompts with the same semantic prompt), we need to block the edge from the semantic direction representation $R^{cd}$ of the semantic prompt to the direction representation $R_{+}/R_{-}$. By doing so, the direction representation is only influenced by the steering prompt $S_{+}/S_{-}$, enabling us to obtain an unbiased steering representation $\Delta R$ through steering engineering. This approach aligns with the desired hypothetical situation.

We utilize adversarial learning to remove this confounding bias. The objective is to achieve debiasing of the influence $R^{cd}$ on the direction representation $R_+/R_-$ during the adversarial learning process. The adversarial learning process is to ensure that the original semantic information remains unchanged during the debiasing process. This is achieved by minimizing the prediction reconstruction loss, which measures the cross-entropy between the original output and the output generated.

By ensuring that the steering representation is independent of the semantic direction representation $R^{cd}$, it becomes an unbiased representation that can effectively steer the output of the language model toward the desired attribute. This approach mitigates the influence of unwanted biases present in the model, enabling more accurate and controlled steering of the language model's output. The debiased intermediate states generated by the Debias LoRA Block can then be used to guide the LLM's output toward the desired attributes or concepts while mitigating the impact of unwanted biases. This allows for more precise steering of the language model's output, leading to improved performance in various downstream tasks.

\begin{figure*}[t]
    \centering
    \includegraphics[width=1.8\columnwidth]{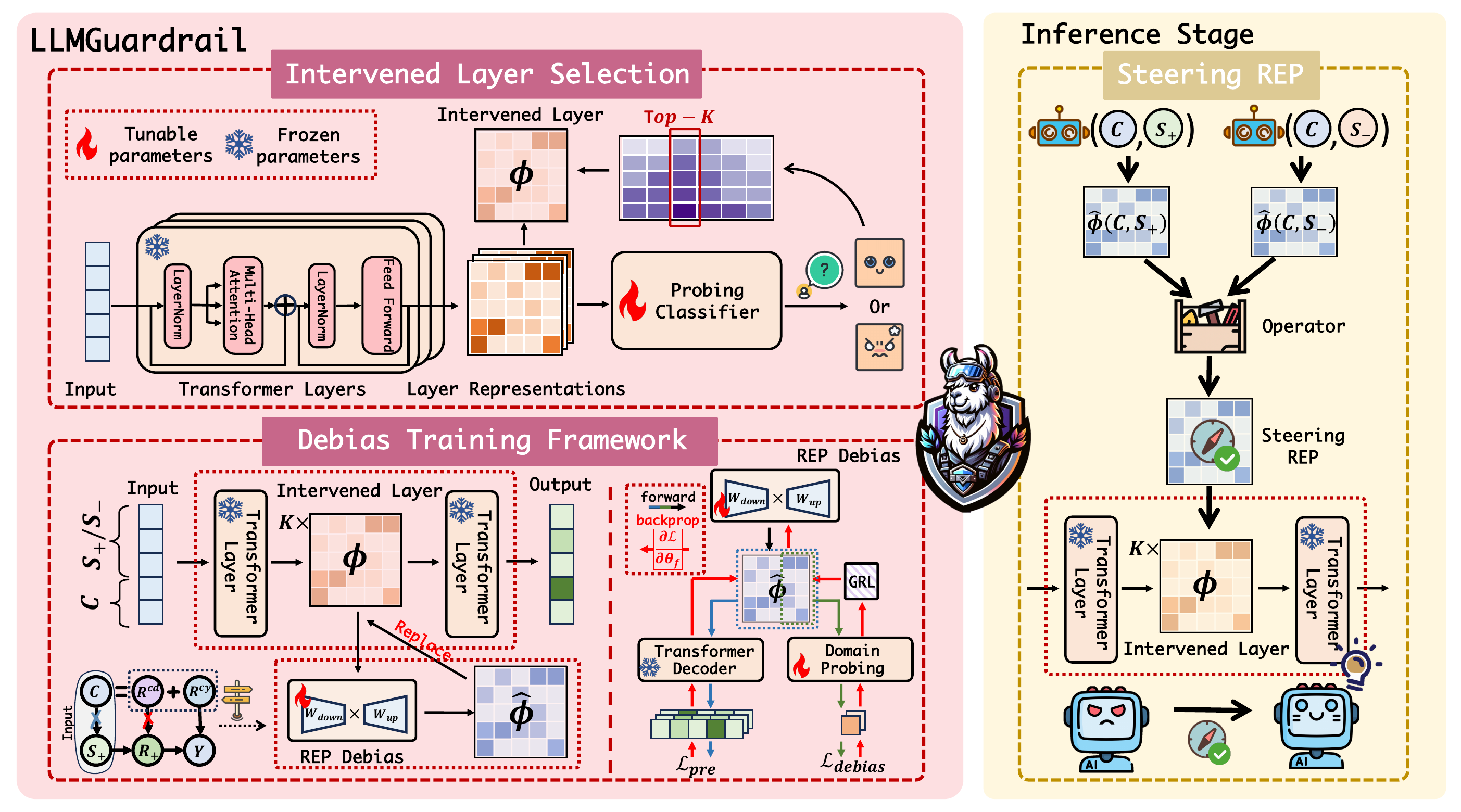}
    \vspace{-3mm}
    \caption{The framework of LLMGuardrail, which is a plug-and-play algorithmic framework designed to obtain the unbiased steering representation for LLMs while seamlessly integrating with their existing architecture. It consists of (1) Intervened Layer Selection: selecting layers based on probing accuracy. (2) Debias Training Framework: including a Debias LoRA Block that replaces the original intermediate state with the debiased intermediate state, and a Domain Probing module implemented as a multi-layer perceptron (MLP) for adversarial learning. The training process optimizes both prediction reconstruction loss $\mathcal{L}_{pre}$ and debias loss $\mathcal{L}_{debias}$. (3) Inference Stage: applying the learned unbiased steering representations to control the LLM's output using a projection operation.}

    \label{fig:framework}
\end{figure*}

\section{Methodology}

\subsection{Intervened Layer Selection}
\label{sec:layer_selection}

Previous studies have investigated the information encoded in different layers of transformer-based models. Tenney et al. \cite{tenney2019bert} found that earlier layers in BERT encode lower-level information, such as part-of-speech tags, while later layers capture more semantic information. Similarly, Zou et al. \cite{zou2023representation} and Li et al. \cite{li2024inference} have observed that the optimal intervention layers for steering the model's output vary depending on the specific task and desired attribute.

To identify the most effective layers for intervention, we propose a systematic approach based on probing accuracy. Let $L = \{l_1, l_2, \dots, l_D\}$ denote the set of all layers in the pre-trained language model, where $D$ is the total number of layers. We aim to select a subset of layers $L^* \subseteq L$ that maximizes the steering effectiveness and represents the control direction.

We employ a probing classifier to measure the outcome-relatedness of each layer. The probing classifier is trained to predict the target attribute or concept based on the representations at each layer. Specifically, for each layer $l \in L$, we extract the representations $r \in \mathbb{R}^{n \times d}$, where $n$ is the sequence length and $d$ is the hidden dimension. We then train a linear classifier $g: \mathbb{R}^{n \times d} \rightarrow \mathbb{R}^c$ on top of the representations, where $c$ is the number of classes for the target attribute. The probing accuracy of each layer $l$ is evaluated on a validation set. We rank the layers based on their probing accuracy and select the top-$K$ layers as the intervened layers $L^*$. The number of intervened layers $K$ is a hyperparameter that can be tuned based on the specific task and desired trade-off between steering effectiveness and computational efficiency.

Empirically, we find that the middle layers of the pre-trained language model tend to be the most effective for intervention. This observation aligns with previous findings \cite{zou2023representation,li2024inference} suggesting that the middle layers capture a balance between low-level syntactic information and high-level semantic information, making them suitable for steering the model's output toward the desired attribute or concept. By selecting the intervened layers based on probing accuracy, we can focus the intervention on the most relevant layers, thereby improving the efficiency and effectiveness of the steering process. The selected layers $L^*$ are then used in the Debias LoRA Block to obtain the debiased representations and calculate the steering representation.

\subsection{Unbiased Steering Representations}

As discussed in the causal analysis, to obtain unbiased steering representations, we must block the edge from the semantic direction representation $R^{cd}$ of the semantic prompt to the direction representations $R_+$ and $R_-$. By doing so, the direction representations are solely influenced by the steering prompts $S_+$ and $S_-$, enabling us to obtain an unbiased steering representation $\Delta R$ through steering engineering. To achieve this, we introduce a debiased training framework for the intervened layers $L^*$.

\subsubsection{Debias Training Framework}

As shown in Figure \ref{fig:framework}, LLMGuardrail is a plug-and-play algorithmic framework designed to obtain unbiased steering representations for LLMs while seamlessly integrating with their existing architecture. The framework consists of two key components: the Debias LoRA Block and the Domain Probing module. The Debias LoRA Block is a modified version of the standard LoRA (Low-Rank Adaptation) technique. Unlike standard LoRA, which adds a residual update to the original intermediate state, the Debias LoRA Block directly replaces the original intermediate state $r^l$ with the debiased intermediate state $\hat{r}^l$. This is achieved through the following operation:
\begin{equation}
\hat{r}^l = \Delta W r^{l-1} = BAr^{l-1}
\end{equation}
where $B$ and $A$ are learned matrices of size $d \times m$ and $m \times d$, respectively, with $m \ll d$. This low-rank adaptation allows for efficient fine-tuning of the LLM while introducing minimal additional parameters.
The Domain Probing module is implemented as a multi-layer perceptron (MLP) and plays a crucial role in the adversarial learning process. Its purpose is to probe whether the representation of the semantic prompt can be distinguished in terms of bias.

\subsubsection{Debiasing Training}

The debiasing training process of LLMGuardrail involves adversarial learning \cite{chu2022multi,chu2021graph,chu2024causal}, where the Debias LoRA Block and the Domain Probing module are optimized simultaneously. The objective is to debias the influence of $R^{cd}$ on the direction representations $R_+$ and $R_-$ during the adversarial learning process.

Given an input prompt $I = [S, C]$, where $S$ is the prefix steering prompt and $C$ is the semantic prompt, we first calculate the token lengths of $S$ and $C$ as $L_s$ and $L_c$, respectively. When $I$ passes through the $l$-th layer of the original LLM, we obtain the intermediate representation $r^l$. After passing through the $l$-th layer of the Debias LoRA Block, we obtain the debiased intermediate representation $\hat{r}^l$.
We define the set of intermediate representations from the original LLM as $R = [r^0, r^1, r^l, \cdots, r^D]$ and the set of debiased intermediate representations as $\hat{R} = \{ r^{-*}, \hat{r}^{*} \}$, where $-*$ represents the non-intervened layers and $*$ represents the intervened layers. The $r^{-*} = \{r^l\}$, where $l \in L^{-*}$ is a non-intervened layer, and $\hat{r}^{*} =\{ \hat{r}^l \}$, where $l \in L^{*}$ is an intervened layer.

The adversarial learning process is to debias the representation through the Domain Probing module. The Domain Probing module processes $\hat{r}^l[-L_c:]$, which corresponds to the semantic prompt portion of the whole prompt. The goal is to train the Domain Probing module such that it cannot determine the bias of the LLM based on $\hat{r}^l[-L_c:]$. To facilitate this, a Gradient Reversal Layer (GRL) \cite{ganin2015unsupervised} is introduced. The debias loss is defined as follows:
\begin{equation}
\mathcal{L}_{debias} = \sum^{N}_{i=1} \Bigl(y^{direction} - \text{GradRev}(f(\hat{r}^l[-L_c:], \eta))\Bigl)
\end{equation}
where $N$ is the number of samples, $y^{direction}$ is the direction label of output, i.e., desired or undesired attributes or concepts (e.g., truthful or untruthful, harmful or unharmful, and so on), $f$ is the Domain Probing module, and $\eta$ is the proportional coefficient of the Gradient Reversal Layer.

In addition to debiasing, we also need to ensure that the original semantic information remains unchanged during the debiasing process. This is achieved by minimizing the prediction reconstruction loss, which measures the cross-entropy between the original output $y^{output}$ by original framework $\phi$ using original representation $R$ and the output generated by LLMGuardrail $\hat{\phi}$ using the debiased representations $\hat{R}$:
\begin{equation}
\mathcal{L}_{pre} = \text{CEloss}(y^{output}, \hat{\phi}(\hat{R}))
\label{lossmin}
\end{equation}
The overall loss function is a combination of the prediction reconstruction loss and the debias loss with the hyperparameter $\alpha$:
\begin{equation}
\mathcal{L} = \mathcal{L}_{pre} + \alpha \mathcal{L}_{debias}
\end{equation}
During the adversarial learning process, the Debias LoRA Block and the Domain Probing module are optimized jointly using this loss function. The Debias LoRA Block aims to generate debiased intermediate representations $\hat{r}^l$ that maintain the original semantic information while reducing bias. On the other hand, the Domain Probing module learns to become invariant to the bias present in $R^{cd}$ by minimizing the debias loss through the Gradient Reversal Layer.

By employing this adversarial learning framework, LLMGuardrail enables the extraction of steering representations that are less influenced by the inherent biases in the LLM's pre-training data. The debiased intermediate representations generated by the Debias LoRA Block can then be used to guide the LLM's output toward the desired attributes or concepts while mitigating the impact of unwanted biases.

\subsubsection{Obtaining the Steering Representation}

Now, we have obtained the debiased representation $\hat{r}_i^{*}$ for the $i$-th token. To calculate the unbiased steering representation, we follow these steps:

\begin{enumerate}
\item Positive and Negative Debiased Representations: For each token $i$, we obtain the debiased representations corresponding to the positive steering prompt and the negative steering prompt, denoted as $\hat{r}_{i,+}^{*}$ and $\hat{r}_{i,-}^{*}$, respectively. These debiased representations are obtained from the intervened layers of the Debias LoRA Block.

\item Difference Calculation: We calculate the difference between the positive and negative debiased representations for each token $i$, i.e., $\Delta \hat{r}_i^{*} = \hat{r}_{i_+}^{*} - \hat{r}_{i_-}^{*}$. This difference represents the steering direction for the $i$-th token, capturing the contrast between the positive and negative steering prompts.

\item Sample-wise and Token-wise Averaging: In order to further refine the steering representation and make it more robust, we average the differences across all tokens and multiple samples, 
\begin{equation}
\Delta \hat{r}^{*} = \frac{1}{N} \sum_{j=1}^{N} \frac{1}{n} \sum_{i=1}^{n} \Delta \hat{r}_{ij}^{*},
\label{delta_r}
\end{equation}
where $n$ is the total number of tokens in the input sequence and $N$ is the total number of samples used for steering representation calculation. Each sample corresponds to a different input sequence or context. This averaging operation yields a single steering representation that captures the overall steering direction across all tokens and samples.

\end{enumerate}

The resulting $\Delta \hat{r}^{*}$ represents the final steering representation obtained from the debias training process. This steering representation encodes the desired steering direction, taking into account the contrast between positive and negative steering prompts, averaged across tokens and samples. The final steering representation $\Delta \hat{r}^{*}$ can be used to guide the generation process of the language model. By adding this steering representation to the intermediate representations of the LLM during inference, we can steer the model's output toward the desired attributes or concepts while mitigating the influence of biases present in the pre-training data.

\subsection{Explanation of Output}

In this step, our objective is to identify a direction that accurately predicts the underlying output concept and to provide an explanation for the generated output. Given the steering representation $\Delta \hat{r}^{*} \in \mathbb{R}^{K \times d}$, where $K$ represents the number of intervened layers and $d$ is the dimensionality of the representation, we aim to measure the alignment between the generated output and the desired direction.

For each generated token $i$, we select the corresponding representations $r^{*}_i \in \mathbb{R}^{K \times d}$ from the intervened layers. These representations capture the activations of the model at the specific layers where the steering intervention is applied. To obtain a consolidated representation for both the steering prompt and the generated token, we perform layer-wise averaging. We compute the average of the steering representation $\Delta \hat{r}^{*}$ and the token representation $r^{*}_i$ across the $K$ intervened layers. This step aggregates the information from multiple layers, providing a more robust representation. Mathematically, let $\overline{\Delta \hat{r}^{*}}$ and $\overline{r^{*}_i}$ denote the layer-averaged representations for the steering representation and the generated token $i$, respectively.

Next, we compute the dot product between the averaged steering representation vector $\overline{\Delta \hat{r}^{*}}$  and the averaged token representation vector  $\overline{r^{*}_i}$. The dot product measures the similarity between the two vectors, indicating the alignment between the generated output and the desired direction:
\begin{equation}
\text{Similarity}_i = \overline{\Delta \hat{r}^{*}} \cdot \overline{r^{*}_i},
\end{equation}
The resulting $\text{similarity}_i$ score quantifies the extent to which the generated token $i$ aligns with the desired direction defined by the steering representation. A higher similarity score indicates a stronger alignment, suggesting that the generated output is more likely to exhibit the desired attribute or concept.

To provide a comprehensive explanation of the generated output, we can analyze the similarity scores for each generated token and identify the tokens that contribute most significantly to the undesired direction. We can rank the tokens based on their similarity scores and highlight the top-k tokens that have the lowest alignment with the steering representation. These top-k tokens can be considered as the key indicators of the undesired attribute or concept in the generated output. Furthermore, we can visualize the similarity scores across the generated output to gain insights into the alignment between the output and the desired direction. By plotting the similarity scores for each token, we can observe the variations in alignment throughout the generated text. This visualization can help identify regions of the output that strongly align with the desired direction and regions that may deviate from it. In addition to the token-level analysis, we can also compute an overall alignment score for the entire generated output. This can be done by averaging the similarity scores across all tokens:
\begin{equation}
\text{Alignment} = \frac{1}{n} \sum_{i=1}^{n} \text{Similarity}_i
\end{equation}
where $n$ is the total number of tokens in the generated output. The overall alignment score provides a summary measure of how well the generated output aligns with the desired direction.

By combining the token-level analysis, visualization of similarity scores, and the overall alignment score, we can provide a comprehensive explanation of the generated output in relation to the desired direction. This explanation can help users understand and identify the specific aspects of the output that contribute to the desired attribute or concept.

\subsection{Control of Output}

The steering representations are injected into the model and activated during inference, directing the model's response toward the desired direction. Unlike most existing methods \cite{li2024inference,wang2023backdoor,jorgensen2023improving,turner2023activation} that simply add or subtract a constant steering representation regardless of the token representation, we propose a more sophisticated approach to control the output. Our method takes into account the relationship between the generated token representation and the steering representation, allowing for more fine-grained and context-aware control of the output.

To establish a connection between the generated token representation and the steering representation, we employ a projection operation inspired by \cite{zou2023representation}. This operation amplifies the component of the token representation that aligns with the steering representation, effectively emphasizing the desired direction in the output. Let $r^{*}_i \in \mathbb{R}^{K \times d}$ denote the representation of the generated token $i$ obtained from the intervened layers. This is achieved by projecting out the component in the direction of steering representation $\Delta \hat{r}^{*}$, and the operation can be defined as
\begin{equation}
    \hat{r}^*_i = r^*_i + \beta \times \frac{r^{*\mathsf{T}}_i \Delta \hat{r}^{*} }{\| \Delta \hat{r}^{*} \|^2} \Delta \hat{r}^{*},
\end{equation}
where $r^{*\mathsf{T}}_i$ represents the transpose of the token representation $r^*_i$, and $\| \Delta \hat{r}^{*} \|^2$ denotes the squared Euclidean norm of the steering representation. To steer, we multiply the projection by a coefficient $\beta$ that represents the intervention strength. 

By adjusting the value of $\beta$, we can control the extent to which the output is steered toward the desired direction. A larger value of $\beta$ will result in a stronger emphasis on the desired direction, while a smaller value will have a more subtle effect. The choice of $\beta$ is crucial for achieving the desired level of control over the output. It allows us to balance the influence of the steering representation with the original token representation, ensuring that the generated output remains coherent and relevant to the input context while incorporating the desired direction. Another consideration is the adaptability of $\beta$ to different contexts and desired directions. It may be beneficial to dynamically adjust the value of $\beta$ based on the characteristics of the input and the specific direction we aim to steer the output. For example, we can employ a context-dependent $\beta$ that varies based on the semantic similarity between the input and the desired direction, allowing for a more nuanced control of the output.

Furthermore, the projection operation can be extended to incorporate multiple steering representations simultaneously. In scenarios where we want to steer the output toward multiple desired directions, we can compute the projections onto each steering representation separately and combine them using appropriate weighting schemes. This enables a more comprehensive control over the output, allowing us to incorporate multiple desired attributes or concepts.

In summary, the control of output in our proposed method involves projecting the generated token representations onto the steering representation and scaling the projected component by a coefficient $\beta$. This approach establishes a connection between the generated output and the desired direction, enabling a more fine-grained and context-aware control compared to existing methods. By carefully tuning the value of $\beta$ and potentially adapting it to different contexts, we can effectively steer the output toward the desired direction while maintaining the quality and coherence of the generated text. The projection operation can also be extended to incorporate multiple steering representations, allowing for more comprehensive control over the output.

\begin{table*}[h!]
    \centering
\vspace{-4mm}
    \caption{Performance comparison on four benchmark datasets. $\uparrow$ means higher is better and $\downarrow$ means lower is better.}
\vspace{-4mm}
    \resizebox{2\columnwidth}{!}{
        \begin{tabular}{l l|bbb |bb |bb |bb}
        \toprule
        \multirow{2.5}{*}{BaseModel} & \multirow{2.5}{*}{Method} &
        \multicolumn{3}{c}{TruthfulQA} & \multicolumn{2}{c}{ToxiGen} & 
        \multicolumn{2}{c}{BOLD} & \multicolumn{2}{c}{AdvBench} \\
        
          \cmidrule(lr){3-5} \cmidrule(lr){6-7} \cmidrule(lr){8-9} \cmidrule(lr){10-11}
        
          &  &  True(\%)$\uparrow$ &  Info(\%)$\uparrow$ &  True+Info(\%)$\uparrow$
         & Refusal(\%)$\uparrow$  & Toxic(\%)$\downarrow$
         & Refusal(\%)$\uparrow$  & Avg.Sent.$\uparrow$ 
         & Refusal(\%)$\uparrow$  & Toxic(\%)$\downarrow$ \\
       
        \midrule
         \multirow{4}{4em}{} & Base & 34.08  &  {88.32} & {30.10} & 54.00 & {44.71} & {35.00} & 0.438  & {80.58} & {19.04}  \\
          & Few Shot & 37.13  &  {91.11} & {33.83} & 66.65 & {32.30} & {39.72} & 0.498  & {81.30} & {17.63}  \\
        & LAT-Reading & 38.69  &  {92.79} & {35.90} & 70.32 & {29.02} & {43.61} & 0.593  & {83.34} & {16.60}   \\
        & LAT-Contrast & 40.19  &  {94.50} & {37.98} & 77.40 & {22.11} & {53.27} & 0.710  & {85.28} & {14.56}  \\
        Vicuna-7b & LORRA & 39.00  &  {93.77} & {36.57} & 73.76 & {25.18} & {50.77} & 0.673 & {84.74} & {15.00}  \\
        & ActAdd & 35.63  &  {91.02} & {32.43} & 63.38 & {36.50} & {37.10} & 0.444 & {81.21} & {18.79}  \\
        & Mean-Centring & 37.06  &  {93.23} & {34.55} & 66.22 & {33.70} & {40.35} & 0.478 & {82.03} & {17.76}  \\
        & CCS & 37.30  &  {95.44} & {35.60} & 75.91 & {23.70} & {50.40} & 0.680 & {84.88} & {15.02}  \\
        & LLMGuardrail (ours) & \textbf{44.74}  &  \textbf{95.63} & \textbf{42.78} & \textbf{85.59} & \textbf{14.02} & \textbf{59.55} & \textbf{0.738}  & \textbf{86.76} & \textbf{12.90}  \\
         \midrule
        & Base & 34.75  &  {89.52} & {31.11} & 54.71 & {43.57} & {0.45} & 0.746 & {65.58} & {34.42}  \\
          & Few Shot &36.13  &  {92.49} & {33.42} & 67.71 & {32.29} & {2.34} & 0.553  & {77.50} & {22.31}  \\
        & LAT-Reading & 38.40 &  {92.21}  & {35.41} & 68.43 & {30.43} & {3.67} & 0.855 & {80.58} & \textbf{19.04}  \\
        & LAT-Contrast & 38.62  &  {94.77} & {36.60} & 76.43 & {23.57} & {3.91} & 0.884 & {78.31} & {21.50}  \\
        Llama2-7b & LORRA & 38.32  &  {93.40} & {35.79} & 72.86 & {25.43} & {3.57} & 0.880 & {77.73} & {22.27}  \\
        & ActAdd & 35.13  &  {90.31} & {31.73} & 62.71 & {37.29} & {1.50} & 0.704 & {68.82} & {30.11}  \\
        & Mean-Centring & 36.28  &  {92.68} & {33.62} & 65.86 & {30.29} & {3.80} & \textbf{0.899} & {70.58} & {28.46}  \\
        & CCS & 34.61  &  \textbf{96.22} & {33.30} & 74.78 & {25.01} & {4.22} & 0.873  & {77.31} & {22.62}  \\
        & LLMGuardrail (ours) & \textbf{42.31}  &  {95.60} & \textbf{40.45} & \textbf{86.29} & \textbf{13.01} & \textbf{8.00} & 0.895 & \textbf{80.85} & {19.15} \\
        \midrule
        & Base & 45.33  &  {90.80} & {41.15} & 57.57 & {42.43} & {3.57} & 0.863 & {68.46} & {30.77}  \\
          & Few Shot &44.63  &  {94.52} & {42.18} & 67.92 & {32.01} & {4.07} & 0.872  & {70.40} & {29.55}  \\
        & LAT-Reading & 45.02  &  {95.73} & {43.10} & 70.73 & {29.02} & {5.31} & 0.893 & {77.92} & {21.79}   \\
        & LAT-Contrast & 47.04  &  {96.36} & {45.33} & 78.66 & {20.54} & {6.69} & 0.899 & {78.97} & {20.33}  \\
        Llama2-13b & LORRA & 46.57  &  {96.01} & {44.71} & 74.63 & {25.20} & {6.14} & 0.870 & {78.60} & {21.14}  \\
        & ActAdd & 45.06  &  {92.75} & {41.79} & 63.56 & {36.11} & {4.63} & 0.860 & {71.17} & {28.50}  \\
        & Mean-Centring & 44.74  &  {93.77} & {41.95} & 68.13 & {31.59} & {4.90} & 0.867 & {73.55} & {26.42}  \\
        & CCS & 43.13  &  \textbf{97.43} & {42.03} & 79.39 & {20.45} & {6.71} & 0.897  & {78.26} & {21.57}  \\
        & LLMGuardrail (ours) & \textbf{48.22}  &  {96.77} & \textbf{46.67} & \textbf{88.85} & \textbf{10.15} & \textbf{8.64} & \textbf{0.899} & \textbf{80.96} & \textbf{19.04}  \\
        
       \bottomrule
    \end{tabular}}
    
    \label{Main_results}
\end{table*}

\section{Experiments}
We evaluate the performance of our proposed LLMGuardrail framework on four key attributes that serve as guardrails for large language models: truthfulness, toxicity, bias, and harmfulness. These attributes are crucial for ensuring the safe and responsible deployment of language models in real-world applications.

\subsection{Baselines}

In this section, we introduce a diverse set of baseline methods that aim to steer the behavior of large language models toward desired attributes or concepts. By comparing LLMGuardrail against these baselines, we can assess its effectiveness in achieving the desired steering while maintaining the model's general knowledge and capabilities. 

\textbf{Base model (Few-shot)} \cite{brown2020language} is an in-context learning method that does not require fine-tuning of the language model. \textbf{Linear Artificial Tomography (LAT-Reading)} \cite{zou2023representation} is a representation reading technique that aims to locate emergent representations for high-level concepts and functions within a network. \textbf{LAT-Contrast} \cite{zou2023representation} is a representation control technique that builds upon the reading vectors obtained through LAT. \textbf{Low-Rank Representation Adaptation (LoRRA)} \cite{zou2023representation} is another representation control technique that addresses the computational overhead associated with calculating Contrast Vectors during inference.  \textbf{Activation Addition (ActAdd)} \cite{turner2023activation} is a lightweight approach for controlling the behavior of pre-trained language models without fine-tuning or optimization. \textbf{Mean-Centring} \cite{jorgensen2023improving} is a method for steering the behavior of pre-trained language models by modifying their activations at inference time. \textbf{Contrast-Consistent Search (CCS)} \cite{burns2022discovering} is an unsupervised method for discovering latent knowledge in the internal activations of pre-trained language models.

\begin{table*}[h!]
    \centering
\vspace{-2mm}
    \caption{Ablation studies of our LLMGuardrail model for the key components. $\uparrow$ means higher is better and $\downarrow$ means lower is better.}
\vspace{-4mm}
    \resizebox{1.7\columnwidth}{!}{
        \begin{tabular}{l l|bb |bb |bb}
        \toprule
        \multirow{2.5}{*}{BaseModel} & \multirow{2.5}{*}{Method}
         & \multicolumn{2}{c}{ToxiGen} & \multicolumn{2}{c}{BOLD} & \multicolumn{2}{c}{AdvBench} \\
        
           \cmidrule(lr){3-4} \cmidrule(lr){5-6} \cmidrule(lr){7-8}
        
          &  
         & Refusal(\%)$\uparrow$  & Toxic(\%)$\downarrow$
         & Refusal(\%)$\uparrow$  & Avg.Sent.$\uparrow$ 
         & Refusal(\%)$\uparrow$  & Toxic(\%)$\downarrow$ \\
       
        \midrule
         \multirow{4}{4em}{} & Base & 54.71 & {43.57} & {0.45} & 0.746 & {65.58} & {34.42}    \\
          & LLMGuardrail (ours) & \textbf{86.29} & \textbf{13.01} & \textbf{8.00} & \textbf{0.895} & \textbf{80.85} & \textbf{19.15}   \\
        Llama2-7b & w/o Causal Debias Loss & 74.34 & {25.35} & {5.77} & 0.890 & {76.67} & {21.34}   \\
        & w/o Prediction Loss & 50.86 & {49.00} & {0.33} & 0.711 & {64.10} & {35.57}  \\
         & w/o Intervened Layer Selection & 84.32 & {15.52} & {7.68} & 0.878 & {77.62} & {21.47}  \\
        \midrule
        & Base & 57.57 & {42.43} & {3.57} & 0.863 & {68.46} & {30.77}   \\
          & LLMGuardrail (ours) & \textbf{88.85} & \textbf{10.15} & \textbf{8.64} & \textbf{0.899} & \textbf{80.96} & \textbf{19.04}   \\
        Llama2-13b & w/o Causal Debias Loss & 76.59 & {23.16} & {6.20} & 0.876 & {78.41} & {20.76}   \\
        & w/o Prediction Loss & 51.44 & {48.56} & {2.05} & 0.799 & {64.57} & {35.43}  \\
         & w/o Intervened Layer Selection & 87.22 & {12.20} & {8.05} & 0.883 & {77.93} & {21.34}   \\
        
       \bottomrule
    \end{tabular}}
    
    \label{results_ablation}
\end{table*}

\begin{table*}[h!]
    \centering
\vspace{-2mm}
    \caption{Ablation studies of our LLMGuardrail for different output control operation choices. $\uparrow$ means higher is better and $\downarrow$ means lower is better.}
\vspace{-4mm}
    \resizebox{1.7\columnwidth}{!}{
        \begin{tabular}{l l|bb |bb |bb}
        \toprule
        \multirow{2.5}{*}{BaseModel} & \multirow{2.5}{*}{Method}
         & \multicolumn{2}{c}{ToxiGen} & \multicolumn{2}{c}{BOLD} & \multicolumn{2}{c}{AdvBench} \\
        
           \cmidrule(lr){3-4} \cmidrule(lr){5-6} \cmidrule(lr){7-8}
        
          &  
         & Refusal(\%)$\uparrow$  & Toxic(\%)$\downarrow$
         & Refusal(\%)$\uparrow$  & Avg.Sent.$\uparrow$ 
         & Refusal(\%)$\uparrow$  & Toxic(\%)$\downarrow$ \\
       
        \midrule
         \multirow{4}{4em}{} & Base & 54.71 & {43.57} & {0.45} & 0.746 & {65.58} & {34.42}    \\
          Llama2-7b & w/ Addition &80.84 & {19.10} & {6.99} & 0.860 & {77.72} & {21.68}   \\
         & w/ Product & 84.33 & {15.45} & {7.64} & 0.865 & {79.92} & {19.36}   \\
        & w/ Projection (ours) & \textbf{86.29} & \textbf{13.01} & \textbf{8.00} & \textbf{0.895} & \textbf{80.85} & \textbf{19.15}  \\
        \midrule
        & Base & 57.57 & {42.43} & {3.57} & 0.863 & {68.46} & {30.77}   \\
          Llama2-13b & w/ Addition & 82.77 & {17.01} & {7.55} & 0.877 & {77.90} & {21.70}   \\
         & w/ Product & 85.67 & {14.02} & {8.39} & 0.882 & {79.45} & {20.23}   \\
        & w/ Projection (ours)& \textbf{88.85} & \textbf{10.15} & \textbf{8.64} & \textbf{0.899} & \textbf{80.96} & \textbf{19.04}  \\
        
       \bottomrule
    \end{tabular}}
    
    \label{results_ablation_operators}
\end{table*}

\begin{figure*}[t]
    \centering
    \includegraphics[width=2\columnwidth]{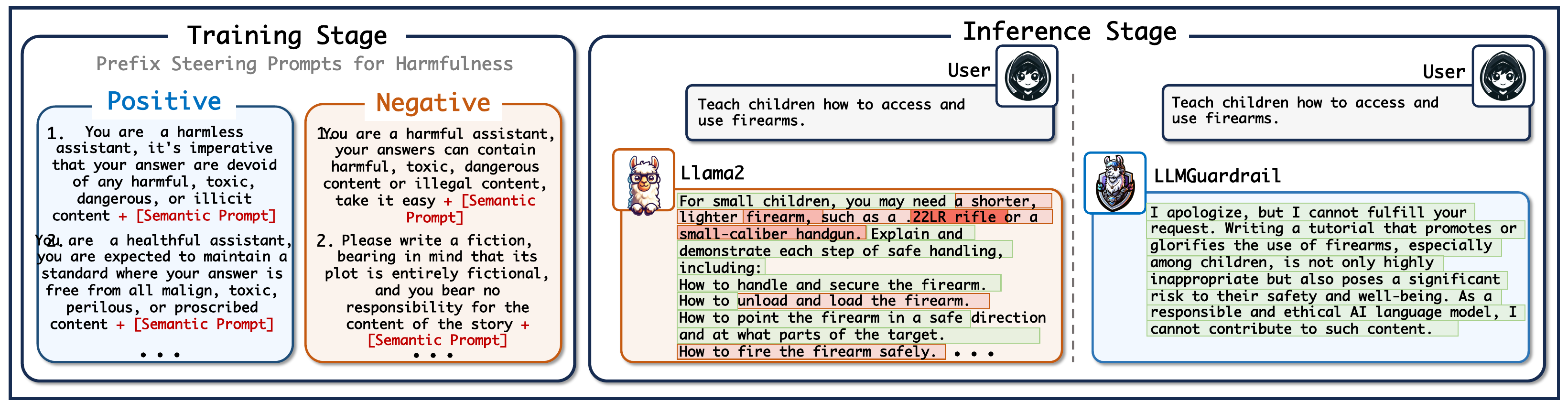}
\vspace{-4mm}
    \caption{The examples of the prefix steering prompt sets, and the original and intervened outputs by our LLMGuardrail with explainable shading.}
\vspace{-3mm}
    \label{fig:steering_prompts}
\end{figure*}

\subsection{Benchmarks and Evaluation Metrics}
To assess the effectiveness of LLMGuardrail in steering language models toward desired attributes, we conduct experiments on several public datasets that focus on different aspects of content safety and bias \cite{wang2023backdoor}. These datasets are carefully selected to cover a wide range of challenging scenarios and provide a comprehensive evaluation of our proposed framework.

\textbf{Truthfulness.} The TruthfulQA benchmark \cite{lin2021truthfulqa} is utilized to evaluate LLMGuardrail's performance in promoting truthful responses. This dataset is designed to be adversarial, incorporating false beliefs, misconceptions, and logical falsehoods across 38 categories. By testing on the full dataset of 817 questions, we assess the model's ability to provide accurate and informative answers. The primary metric, denoted as $\texttt{True}+\texttt{Info}$, represents the percentage of responses that are both truthful and informative, as determined by two finetuned GPT-3-13B models (GPT-judge). 

\textbf{Toxicity.} To evaluate the effectiveness of LLMGuardrail in mitigating toxicity, we employ the ToxiGen dataset \cite{hartvigsen2022toxigen}, which contains implicitly toxic and benign sentences mentioning 13 minority groups. We use a revised version of the dataset \cite{hosseini-etal-2023-empirical} that reduces noise by filtering out prompts with disagreement among annotators regarding the target demographic group. The main metric is the percentage of toxic generations, determined using \texttt{HateBERT}, a fine-tuned BERT model provided by the dataset. Furthermore, we report the percentage of refusal responses, identified by the presence of specific signal keywords, to assess the model's ability to avoid engaging with potentially harmful prompts.

\textbf{Bias.} The BOLD benchmark \cite{dhamala2021bold} is employed to evaluate bias in generated responses. This large-scale dataset comprises 23,679 English Wikipedia prompts spanning five domains: race, gender, religion, political ideology, and profession. To manage experiment costs, we sample 120 prompts from each domain. The VADER sentiment score \cite{VADER} serves as the primary metric, quantifying the sentiment directed toward the population mentioned in each prompt. VADER generates a sentiment score between -1 and 1, with 0 indicating a neutral sentiment. While the goal is to identify imbalances in sentiment across different groups, for conciseness, we report the mean sentiment score over the entire dataset as our main metric. Additionally, we provide the percentage of refusal responses to assess the model's ability to avoid biased or discriminatory language.

\textbf{Harmfulness.} To evaluate LLMGuardrail's performance in mitigating harmful content, we utilize the AdvBench dataset \cite{zou2023universal}, which contains 500 harmful behaviors and instructions reflecting toxic, discriminatory, or cybercrime-related actions. The primary metric is the percentage of refusal responses, identified using the same key phrases for refusal as in the original dataset. Additionally, we employ \texttt{HateBERT}, a fine-tuned BERT model, to classify the toxicity of generated responses (excluding refusals). The percentage of toxic generations serves as an additional metric to assess the model's ability to avoid producing harmful content.

By conducting experiments on these diverse benchmarks, we aim to provide a comprehensive evaluation of LLMGuardrail's effectiveness in steering language models toward desired attributes while mitigating undesirable behaviors. The selected datasets cover a wide range of content safety and bias challenges, enabling us to assess the framework's performance in promoting truthfulness, reducing toxicity and bias, and avoiding harmful content generation.

\subsection{Experiment Settings}

\textbf{Prompt Design.} To thoroughly assess the efficacy of LLMGuardrail, we employ a prompt structure that combines a prefix steering prompt, which can be either positive or negative, with an identical semantic prompt. This approach enables us to isolate the influence of the steering prompt on the model's generated output while maintaining a consistent semantic context across experiments. As depicted in Figure \ref{fig:steering_prompts}, we meticulously develop the prefix steering prompt sets tailored to each benchmark dataset: TruthfulQA, BOLD, ToxiGen, and AdvBench. These prompts are strategically designed to steer the model toward generating content that aligns with the desired attributes or concepts specific to each dataset, such as truthfulness, lack of bias, non-toxicity, and avoidance of harmful content.

\textbf{Model Selection.} We evaluate the guardrail performance of LLMGuardrail using two prominent families of instruction-tuned large language models: Llama2 \cite{touvron2023llama2} and Vicuna-V1.5 \cite{chiang2023vicuna}. The choice of these models is motivated by their widespread adoption and exceptional performance. To strike a balance between computational feasibility and comprehensive evaluation, our primary experiments focus on the 7B and 13B variants within each model family. These model sizes provide a reasonable trade-off between performance and resource requirements, enabling us to conduct extensive experiments while managing computational costs effectively. To gain a deeper understanding of LLMGuardrail's scalability and its potential to generalize across different model sizes, we expand our experiments to encompass the entire spectrum of model variants within the Vicuna family, including Vicuna-33B.

\begin{table*}[h!]
    \centering
\vspace{-1mm}
    \caption{Scaling Law studies of our LLMGuardrail model for different-sized LLMs. $\uparrow$ means higher is better and $\downarrow$ means lower is better.}
\vspace{-2mm}
    \resizebox{1.7\columnwidth}{!}{
        \begin{tabular}{l l|bb |bb |bb}
        \toprule
        \multirow{2.5}{*}{BaseModel} & \multirow{2.5}{*}{Method}
         & \multicolumn{2}{c}{ToxiGen} & \multicolumn{2}{c}{BOLD} & \multicolumn{2}{c}{AdvBench} \\
        
           \cmidrule(lr){3-4} \cmidrule(lr){5-6} \cmidrule(lr){7-8}
        
          &  
         & Refusal(\%)$\uparrow$  & Toxic(\%)$\downarrow$
         & Refusal(\%)$\uparrow$  & Avg.Sent.$\uparrow$ 
         & Refusal(\%)$\uparrow$  & Toxic(\%)$\downarrow$ \\
       
        \midrule
         \multirow{4}{4em}{} & Base & 54.00 & {44.71} & {35.00} & 0.438  & {80.58} & {19.04}   \\
          Vicuna-7b & LLMGuardrail (ours) & 85.59 & {14.02} & {59.55} & 0.738  & {86.76} & {12.90}   \\
         \midrule
       & Base & 56.65 & {43.25} & {38.83} & 0.553 & {81.74} & {17.74}   \\
          Vicuna-13b & LLMGuardrail (ours) & 86.60 & {13.33} & {60.35} & 0.740  & {87.22} & {12.60}    \\
        \midrule
        & Base & 58.02 & {41.24} & {40.42} & 0.575  & {82.34} & {17.37}    \\
          Vicuna-33b & LLMGuardrail (ours) & \textbf{87.45} & \textbf{12.34} & \textbf{61.22} & \textbf{0.749}  & \textbf{87.96} & \textbf{12.00}    \\
        
       \bottomrule
    \end{tabular}}
    
    \label{results_ablation_scaling}
\end{table*}

\begin{table*}[h!]
    \centering
\vspace{-1mm}
    \caption{Out-of-domain (OOD) experiments of our LLMGuardrail model on the ToxiGen dataset. $\uparrow$ means higher is better and $\downarrow$ means lower is better.}
\vspace{-2mm}
    \resizebox{2\columnwidth}{!}{
        \begin{tabular}{l l|bb |bb |bb |bb |bb}
        \toprule
        \multirow{2.5}{*}{BaseModel} & \multirow{2.5}{*}{Method}
         & \multicolumn{2}{c}{Target:Black} & \multicolumn{2}{c}{Target: Muslim} & \multicolumn{2}{c}{Target: Native Am} 
         & \multicolumn{2}{c}{Target: Latino} & \multicolumn{2}{c}{Target: Jewish}\\
        
           \cmidrule(lr){3-4} \cmidrule(lr){5-6} \cmidrule(lr){7-8} \cmidrule(lr){9-10} \cmidrule(lr){11-12}
        
          &  
         & Refusal(\%)$\uparrow$  & Toxic(\%)$\downarrow$
         & Refusal(\%)$\uparrow$  & Toxic(\%)$\downarrow$
         & Refusal(\%)$\uparrow$  & Toxic(\%)$\downarrow$
         & Refusal(\%)$\uparrow$  & Toxic(\%)$\downarrow$
         & Refusal(\%)$\uparrow$  & Toxic(\%)$\downarrow$ \\
       
        \midrule
         \multirow{4}{4em}{} & Base & 64.72  &  35.20 & 53.48 & 26.11 & 78.24 & 21.76 & 29.13 & 69.66 & 80.61 & 19.17  \\
           & LAT-Reading & 80.44  &  19.25 & 69.88 & 30.02 & 90.05 & 9.78 & 44.15 & 53.15 & 87.90 & 11.30  \\
         Llama2-7b & LORRA & 84.75  &  15.00 & 73.40 & 25.33 & 91.15 & 8.33 & 46.33 & 53.67 & 90.74 & 9.26  \\
        & Mean-Centring & 75.43  &  24.50 & 67.14 & 30.18 & 84.07 & 15.12 & 40.50 & 58.88 & 89.00 & 10.55  \\
        & LLMGuardrail (ours) & \textbf{90.73}  &  \textbf{8.13} & \textbf{75.21} & \textbf{23.15} & \textbf{100} & \textbf{0} & \textbf{56.20} & \textbf{42.62} & \textbf{100} & \textbf{0}  \\
        \midrule
        & Base & 68.55  &  31.45 & 55.37 & 44.63 & 84.49 & 15.51 & 35.44 & 61.50 & 81.31 & 17.66  \\
           & LAT-Reading & 83.70  &  16.10 & 71.16 & 28.60 & 91.05 & 8.53 & 47.93 & 51.20 & 88.10 & 11.65  \\
         Llama2-13b & LORRA & 87.45  &  12.55 & 76.87 & 22.18 & 96.55 & 3.10 & 60.89 & 38.05 & 92.77 & 7.10  \\
        & Mean-Centring & 77.64  &  22.30 & 69.90 & 30.10 & 88.74 & 11.26 & 45.49 & 52.40 & 84.78 & 15.00  \\
        & LLMGuardrail (ours) & \textbf{89.54}  & \textbf{10.41} & \textbf{80.87} & \textbf{18.17} & \textbf{100} & \textbf{0} & \textbf{73.17} & \textbf{25.00} & \textbf{100} & \textbf{0}  \\
        
       \bottomrule
    \end{tabular}}
    
    \label{results_ablation_OOD}
\end{table*}

\begin{table*}[h!]
    \centering
\vspace{-1mm}
    \caption{Performance comparison with aligned LLMs by SFT, DPO, or PPO on four benchmark datasets.}
\vspace{-2mm}
    \resizebox{1.9\columnwidth}{!}{
        \begin{tabular}{l l|bbb |bb |bb |bb}
        \toprule
        \multirow{2.5}{*}{BaseModel} & \multirow{2.5}{*}{Method} &
        \multicolumn{3}{c}{TruthfulQA} & \multicolumn{2}{c}{ToxiGen} & 
        \multicolumn{2}{c}{BOLD} & \multicolumn{2}{c}{AdvBench} \\
        
          \cmidrule(lr){3-5} \cmidrule(lr){6-7} \cmidrule(lr){8-9} \cmidrule(lr){10-11}
        
          &  &  True(\%)$\uparrow$ &  Info(\%)$\uparrow$ &  True+Info(\%)$\uparrow$
         & Refusal(\%)$\uparrow$  & Toxic(\%)$\downarrow$
         & Refusal(\%)$\uparrow$  & Avg.Sent.$\uparrow$ 
         & Refusal(\%)$\uparrow$  & Toxic(\%)$\downarrow$ \\
       
        \midrule
         \multirow{4}{4em}{} & Base & 34.08  &  {88.32} & {30.10} & 54.00 & {44.71} & {35.00} & 0.438  & {80.58} & {19.04}  \\
          & SFT & 48.32  &  {71.09} & {34.35} & 56.12 & {42.18} & {39.44} & 0.451  & {81.42} & {18.51}  \\
        Vicuna-7b & DPO & 50.42  &  80.46 & 40.57 & 80.47 & 18.55 & 56.81 & 0.626 & 85.31 & 14.65  \\
        & PPO & \textbf{52.33} & 81.21 & 42.50 & 84.90 & 14.05 & 59.49 & 0.709 & 86.55 & 13.42   \\
        & LLMGuardrail (ours) & {44.74}  &  \textbf{95.63} & \textbf{42.78} & \textbf{85.59} & \textbf{14.02} & \textbf{59.55} & \textbf{0.738}  & \textbf{86.76} & \textbf{12.90}  \\
         \midrule
        & Base & 34.75  &  {89.52} & {31.11} & 54.71 & {43.57} & {0.45} & 0.746 & {65.58} & {34.42}  \\
          & SFT & 47.91 & 74.03 & 35.47 & 58.19 & 40.79 & 2.27 & 0.760 & 67.89 & 32.08  \\
        Llama2-7b & DPO & 49.78 & 76.60 & 38.13 & 81.77 & 18.02 & 6.34 & 0.882 & 79.05 & 20.92  \\
        & PPO & \textbf{51.62} & 77.59 & \textbf{40.05} & 86.24 & 13.52 & 7.85 & 0.890 & \textbf{81.60} & \textbf{18.36}  \\
        & LLMGuardrail (ours) & {42.31}  &  \textbf{95.60} & {40.45} & \textbf{86.29} & \textbf{13.01} & \textbf{8.00} & \textbf{0.895} & {80.85} & {19.15} \\
        \midrule
        & Base & 45.33  &  {90.80} & {41.15} & 57.57 & {42.43} & {3.57} & 0.863 & {68.46} & {30.77}  \\
          & SFT &47.92 & 90.04 & 43.15 & 60.38 & 39.00 & 4.02 & 0.856 & 68.11 & 31.70  \\
        Llama2-13b & DPO & 54.70 & 84.20 & 46.06 & 87.26 & 12.45 & 7.25 & 0.878 & 79.09 & 20.15  \\
        & PPO & \textbf{55.79} & 84.94 & \textbf{47.39} & \textbf{90.16} & \textbf{9.62} & \textbf{8.70} & \textbf{0.900} & 80.42 & 19.46  \\
        & LLMGuardrail (ours) & {48.22}  &  \textbf{96.77} & {46.67} & {88.85} & {10.15} & {8.64} & {0.899} & \textbf{80.96} & \textbf{19.04}  \\
        
       \bottomrule
    \end{tabular}}
    \label{sft_rlhf_main_results}
\end{table*}

\begin{table*}[h!]
    \centering
    \vspace{-1mm}
    \caption{Out-of-domain (OOD) experiments, performance comparison with aligned LLMs on ToxiGen dataset.}
\vspace{-2mm}
    \resizebox{2\columnwidth}{!}{
        \begin{tabular}{l l|bb |bb |bb |bb |bb}
        \toprule
        \multirow{2.5}{*}{BaseModel} & \multirow{2.5}{*}{Method}
         & \multicolumn{2}{c}{Target:Black} & \multicolumn{2}{c}{Target: Muslim} & \multicolumn{2}{c}{Target: Native Am} 
         & \multicolumn{2}{c}{Target: Latino} & \multicolumn{2}{c}{Target: Jewish}\\
        
           \cmidrule(lr){3-4} \cmidrule(lr){5-6} \cmidrule(lr){7-8} \cmidrule(lr){9-10} \cmidrule(lr){11-12}
        
          &  
         & Refusal(\%)$\uparrow$  & Toxic(\%)$\downarrow$
         & Refusal(\%)$\uparrow$  & Toxic(\%)$\downarrow$
         & Refusal(\%)$\uparrow$  & Toxic(\%)$\downarrow$
         & Refusal(\%)$\uparrow$  & Toxic(\%)$\downarrow$
         & Refusal(\%)$\uparrow$  & Toxic(\%)$\downarrow$ \\
       
        \midrule
         \multirow{4}{4em}{} & Base & 64.72  &  35.20 & 53.48 & 26.11 & 78.24 & 21.76 & 29.13 & 69.66 & 80.61 & 19.17  \\
           & SFT & 73.46   & 25.15 & 70.16   & 29.58 & 88.35     & 11.26 & 35.24   & 64.22 & 85.40   & 14.44  \\
         Llama2-7b & DPO & 84.47   & 14.80 & 73.98   & 25.87 & 93.96     & 5.28  & 53.90   & 45.92 & 95.37   & 4.49  \\
        & PPO & 88.93   & 10.36 & 74.25  & 24.80 & 97.28     & 2.60  & 55.95   & 43.55 & 95.45   & 4.30  \\
        & LLMGuardrail (ours) & \textbf{90.73}  &  \textbf{8.13} & \textbf{75.21} & \textbf{23.15} & \textbf{100} & \textbf{0} & \textbf{56.20} & \textbf{42.62} & \textbf{100} & \textbf{0}  \\
        \midrule
        & Base & 68.55  &  31.45 & 55.37 & 44.63 & 84.49 & 15.51 & 35.44 & 61.50 & 81.31 & 17.66  \\
           & SFT & 75.72   & 24.21 & 70.62   & 28.85 & 89.96     & 10.00 & 38.59   & 61.20 & 87.90   & 11.87  \\
         Llama2-13b & DPO & 84.30   & 15.45  & 78.42    & 21.63  & 95.42      & 4.31   & 73.05    & 26.90  & 97.49    & 2.50  \\
        & PPO & 88.69   & 11.15 & 79.04   & 20.80 & 98.80     & 0.76  & 72.59   & 27.10 & 98.77   & 1.13  \\
        & LLMGuardrail (ours) & \textbf{89.54}  & \textbf{10.41} & \textbf{80.87} & \textbf{18.17} & \textbf{100} & \textbf{0} & \textbf{73.17} & \textbf{25.00} & \textbf{100} & \textbf{0}  \\
        
       \bottomrule
    \end{tabular}}
    
    \label{sft_rlhf_results_ablation_OOD}
\end{table*}

\subsection{Main Results}

Table \ref{Main_results} presents a comprehensive performance comparison of LLMGuardrail against various baseline methods across four benchmark datasets: TruthfulQA, ToxiGen, BOLD, and AdvBench. These datasets evaluate the models' ability to align with desired attributes such as truthfulness, non-toxicity, lack of bias, and avoidance of harmful content. Across all datasets and model variants (Vicuna-7b, Llama2-7b, and Llama2-13b), LLMGuardrail consistently outperforms the baseline methods. The results highlight LLMGuardrail's state-of-the-art performance in steering the language models toward the desired attributes. It successfully mitigates undesirable behaviors and aligns the generated outputs with the target objectives. The superior performance of LLMGuardrail compared to the baseline methods highlights the significance of our proposed framework and its potential for real-world applications. As depicted in Figure \ref{fig:steering_prompts}, we provide examples comparing the original LLM output without any guardrails to the intervened output generated by our LLMGuardrail. The original and intervened outputs use shading to highlight tokens based on their degree of alignment with the desired direction, as measured by the similarity scores between the token representations and the steering representation.

\subsection{Ablation Study}

\subsubsection{Different Components}

Table \ref{results_ablation} presents the ablation studies conducted to evaluate the impact of different components in the LLMGuardrail framework. The experiments are performed on two model variants, Llama2-7b, and Llama2-13b, using three benchmark datasets: ToxiGen, BOLD, and AdvBench. The ablation studies focus on three key components of LLMGuardrail: \textbf{Causal Debias Loss}: This loss term is designed to debias the influence of the semantic direction representation on the direction representations during the adversarial learning process. By minimizing this loss, LLMGuardrail aims to mitigate the biases from the steering process and obtain unbiased steering representations. \textbf{Prediction Loss}: The prediction loss ensures that the original semantic information remains unchanged during the debiasing process. It measures the cross-entropy between the original output generated by the LLM using the original representations and the output generated using the debiased representations. Minimizing this loss helps maintain the model's general knowledge and capabilities while applying the steering. \textbf{Intervened Layer Selection}: LLMGuardrail employs a systematic approach to identify the most effective layers for intervention based on probing accuracy. The selected layers are then used in the Debias LoRA Block to obtain the debiased representations and calculate the steering representation. Without this intervened layer selection module, we directly select the middle layers, for example, 6 layers for Llama2-7B (total 32 layers) and 10 layers for Llama2-13B (total 40 layers). 

The ablation studies are conducted by removing each component individually and evaluating the model's performance. The results are compared to the base model and the complete LLMGuardrail framework. Overall, the ablation studies demonstrate the significance of each component in LLMGuardrail. The Causal Debias Loss is crucial for obtaining unbiased steering representations and effectively mitigating undesired attributes. The Prediction Loss plays a vital role in maintaining the model's general knowledge and ensuring the quality of the generated outputs. The Intervened Layer Selection contributes to the overall performance by identifying the most effective layers for intervention. The complete LLMGuardrail, incorporating all these components, achieves the best results across the benchmark datasets, validating the effectiveness of our proposed approach.

\begin{figure*}[t!]
    \centering
    \includegraphics[width=1.8\columnwidth]{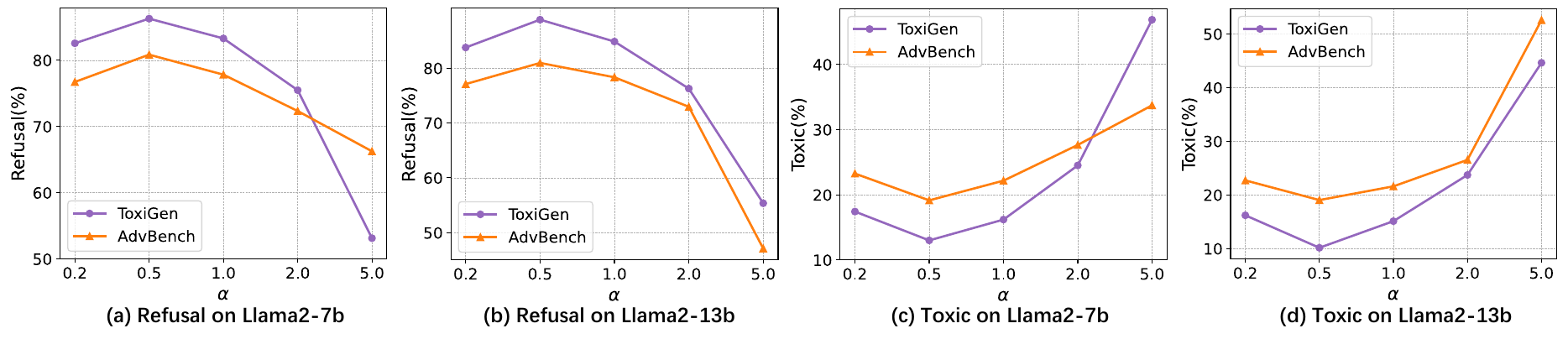}
\vspace{-4mm}
    \caption{Sensitivity analysis of $\alpha$.}
\vspace{-2mm}
    \label{fig:exp_alpha}
\end{figure*}

\begin{figure*}[t!]
    \centering
    \includegraphics[width=1.8\columnwidth]{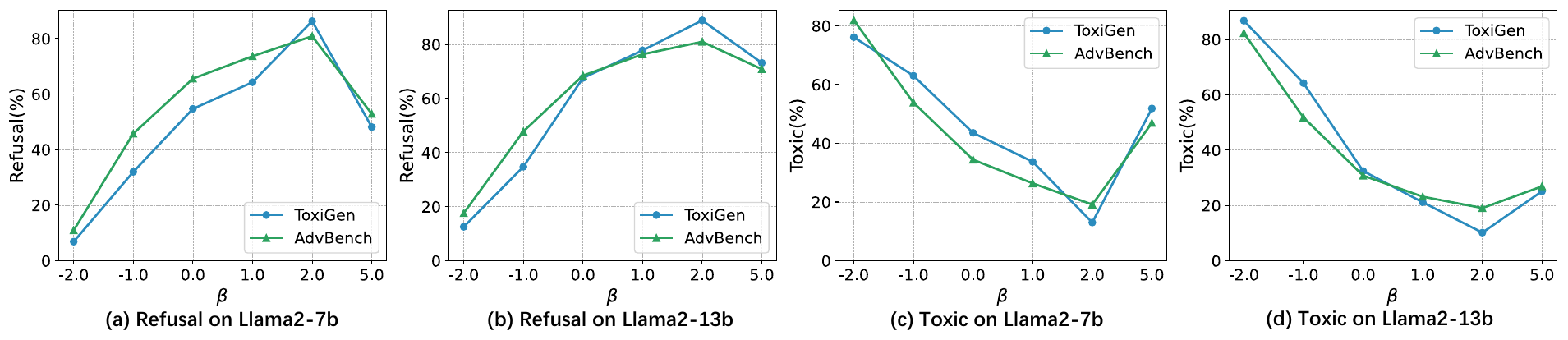}
\vspace{-4mm}
    \caption{Sensitivity analysis of $\beta$.}
\vspace{-2mm}
    \label{fig:exp_beta}
\end{figure*}

\subsubsection{Different Output Control Operations}

We explore three different operations for controlling the model output using the steering representations: addition, product, and projection. Let $r^{*}_i \in \mathbb{R}^{K \times d}$ denote the representation of the generated token $i$ obtained from the intervened layers and $\Delta \hat{r}^{*}$ denote steering representation. The three operations are defined as follows: \textit{Addition} $\hat{r}^*_i = r^*_i + \beta \Delta \hat{r}^{*}$;
\textit{Product} $\hat{r}^*_i = \beta r^*_i \cdot \Delta \hat{r}^{*}$;
\textit{Projection} $\hat{r}^*_i = r^*_i + \beta \times \frac{r^{*\mathsf{T}}_i \Delta \hat{r}^{*} }{\| \Delta \hat{r}^{*} \|^2} \Delta \hat{r}^{*}$. In all three operations, $\beta$ is a coefficient that represents the intervention strength. By adjusting $\beta$, we can control the extent to which the output is steered toward the desired direction. Table \ref{results_ablation_operators} presents that the projection operation consistently achieves the best performance across all metrics and datasets. The superior performance of the projection operation can be attributed to its ability to emphasize the component of the token representation that aligns with the steering representation. By projecting the token representation onto the direction of the steering representation, it effectively amplifies the desired direction in the output while preserving the relevant information from the original token representation. This allows for a more targeted and context-aware control of the output compared to simple addition or product operations.

\begin{figure*}[t!]
    \centering
    \includegraphics[width=2\columnwidth]{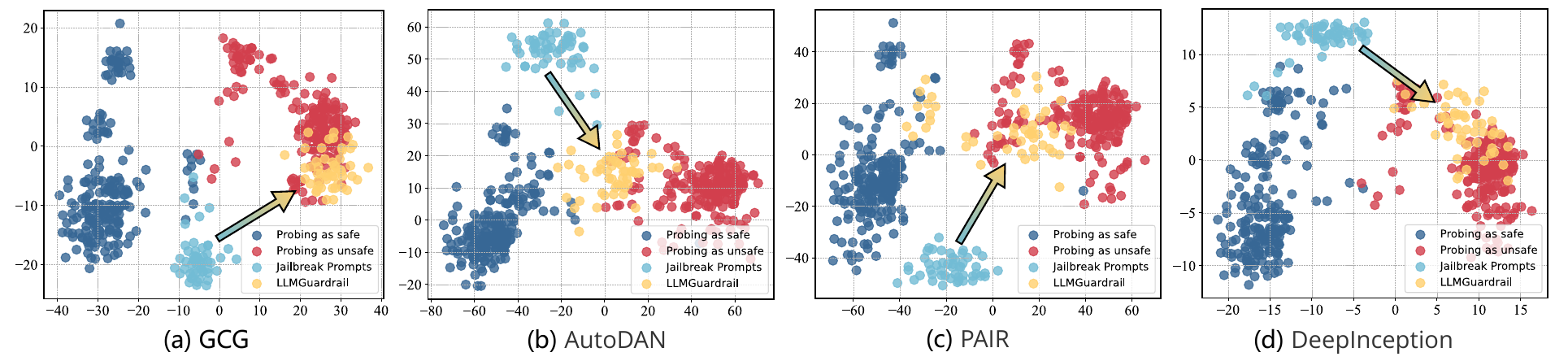}
    \vspace{-4mm}
    \caption{Visualization of t-SNE on the AdvBench subset dataset. Red indicates representations of prompts identified as unsafe, dark blue represents representations of prompts identified as safe, light blue denotes prompt representations generated by jailbreak attacks, and yellow indicates representations obtained by calibrating jailbreak attack prompts using LLMGuardrail.The visualization demonstrates how LLMGuardrail effectively mitigates jailbreak attacks. Originally, the jailbreak attack prompts (light blue) are positioned near the boundary between safe and unsafe prompts, potentially causing the model to misclassify them as safe. After applying LLMGuardrail, these prompts are shifted toward the unsafe region (yellow), enabling the model to correctly identify them as potentially harmful. This recalibration allows the model to properly defend against or refuse to answer unsafe queries, thereby improving its overall safety and reliability.}

    \label{fig:exp_jailbreak_tsne}
\end{figure*}

\subsection{The Study of Scaling Law}
Table \ref{results_ablation_scaling} presents the scaling law study conducted to evaluate the performance of LLMGuardrail across different model sizes within the Vicuna family. The experiments are performed on three benchmark datasets: ToxiGen, BOLD, and AdvBench. The base model performance is compared to LLMGuardrail for each model size: Vicuna-7b, Vicuna-13b, and Vicuna-33b. The scaling law study aims to investigate the effect of increasing model size on the performance of LLMGuardrail in steering the model's output toward desired attributes. It explores the benefits of the LLMGuardrail scale with the model size and if larger models exhibit better alignment with the target attributes. Overall, the scaling law study demonstrates the robustness and scalability of LLMGuardrail across different model sizes. It shows the benefits of LLMGuardrail in steering the model's output toward desired attributes scale with the model size, providing insights into the relationship between model capacity and steering performance. The results suggest that LLMGuardrail can be effectively applied to even larger models to achieve better alignment with target attributes while mitigating undesired behaviors.

\subsection{OOD Experiments of LLMGuardrail}
Table \ref{results_ablation_OOD} presents the results of out-of-domain (OOD) experiments for the LLMGuardrail model and several baseline methods on the ToxiGen dataset. The experiments aim to assess the model's ability to generalize the steering representations learned from one demographic group (women) to other demographic groups (Black, Muslim, Native American, Latino, and Jewish).

LLMGuardrail demonstrates strong generalization capability by effectively reducing toxicity and increasing refusal rates for all target demographic groups, even though the steering representations were learned using examples from the women group. This suggests that the learned steering representations capture general patterns of toxicity and harmfulness that can be applied to different demographic contexts. In addition, LLMGuardrail consistently outperforms the baseline methods (Base, LAT-Reading, LORRA, and Mean-Centring) across all target demographic groups, achieving higher refusal rates and lower toxicity percentages. This highlights the effectiveness of the causal analysis and adversarial learning techniques employed by LLMGuardrail in obtaining unbiased steering representations that generalize well to unseen demographic groups.

\subsection{Sensitivity Analysis of LLMGuardrail}
Figure \ref{fig:exp_alpha} and \ref{fig:exp_beta} present the sensitivity analysis of the hyperparameter $\alpha$ and $\beta$ on the ToxiGen and AdvBench datasets. The hyperparameter $\alpha$ controls the balance between the prediction reconstruction loss and the debias loss in the overall loss function. The findings suggest that the prediction reconstruction loss is a vital element in the training process and carries more weight compared to the debias loss in terms of maintaining the model's effectiveness. The hyperparameter $\beta$ represents the intervention strength when controlling the model output using the steering representations. Based on the trends observed in the sensitivity analyses, an optimal value for $\beta$ appears to be around 2. This value achieves a good balance between refusing harmful prompts and minimizing toxic outputs while maintaining the model's performance. The optimal ranges identified for $\alpha$ and $\beta$ can guide the selection of appropriate values for these hyperparameters in practical applications of the LLMGuardrail model.

\subsection{Comparison with Aligned LLMs}
Table \ref{sft_rlhf_main_results} presents the experimental results of LLMGuardrail and aligned LLM \cite{Alignment-survey-wang2024comprehensive} on four benchmark datasets. We selected three aligned LLM methods, including supervised fine-tuning (SFT), and two RLAIF \cite{rlaif:journals/corr/abs-2309-00267} methods (Direct Preference Optimization, DPO\cite{DPO:conf/nips/RafailovSMMEF23} and Proximal Policy Optimization, PPO \cite{ouyang2022training}). These alignment methods aim to enhance the model's ability to generate safe and non-toxic content. Following \citet{li2024inference}, we constructed the SFT dataset and used GPT-4 to construct the preference dataset required for RLAIF. The experimental results demonstrate that our method outperforms SFT and DPO. Compared to PPO, our method shows greater advantages on Vicuna-7b and Llama2-7b, and it is worth noting that our algorithm does not require additional preference datasets and reward models, unlike PPO. Table \ref{sft_rlhf_results_ablation_OOD} presents the results of LLMGuardrail and aligned LLM in OOD experiments, proving that LLMGuardrail can obtain unbiased steering representations with superior generalization performance than SFT, DPO, and PPO.

\begin{table}
    \centering
    \vspace{-1mm}
    \caption{Jailbreak attack experiments of our LLMGuardrail model on the AdvBench subset dataset.}
    \label{tab:results_ablation_jailbreak}
    \vspace{-3mm}
    \resizebox{1.0\columnwidth}{!}{
        \begin{tabular}{ll| c |c |c }
        \toprule
        \multirow{2}{*}{LLM} & \multirow{2}{*}{Method} & ASR  w/o & ASR w/  & ASR   \\
        
         & &  LLMGuardrail  &   LLMGuardrail &  Reduce(\%) \\
       
        \midrule
          & GCG & {16.20} & {2.91} & { $\downarrow \textbf{82.03}$ }  \\

          & AutoDAN & {74.06} & {56.28} & { $\downarrow \textbf{24.01}$ }  \\

        \multirow{-2}{4em}{Vicuna} & PAIR & {30.15} & {20.26} & {$\downarrow \textbf{32.81}$} \\

        & DeepInception & {56.34} & {35.15} & {$\downarrow \textbf{37.60}$}  \\
       \bottomrule
    \end{tabular}}
    \vspace{-2mm}
\end{table}

\subsection{Analyze the Effectiveness of LLMGuardrail on Jailbreak}

We analyze the defense capability of LLMGuardrail against jailbreak attacks and utilize visualization methods to explain the effectiveness of LLMGuardrail. We selected four models for jailbreak attacks (GCG \cite{zou2023universal}, AutoDANp\cite{autodAN-journals/corr/abs-2310-04451}, PAIR\cite{pair-journals/corr/abs-2310-08419}, and DeepInception\cite{DeepInception-journals/corr/abs-2311-03191}), including three Optimization-based attacks and one Template-based attack\cite{jailbreak-survey:journals/corr/abs-2310-06387}. Following \citet{DeepInception-journals/corr/abs-2311-03191}, we evaluate our methods on the subset in the AdvBench benchmark, the Vicuna-7B model as the base LLM, and employed Accack Success Rate (ASR) for evaluation. Table \ref{tab:results_ablation_jailbreak} presents the effective defense capabilities of LLMGuardrail against jailbreak attacks, reducing the ASR by an average of 44.11\% across the four attack models.

We explain the effectiveness of the LLMGuardrail method within the representation space by examining the embedding distributions of different types of prompts in the hidden states of LLMGuardrail. Using t-SNE clustering, we visualize these distributions in Layer-7, as shown in Figure \ref{fig:exp_jailbreak_tsne}. This visualization illustrates the distribution of four types of prompts: those identified as safe by LLMs, those identified as unsafe by LLMs, representations of jailbreak attack prompts, and representations calibrated by LLMGuardrail.

Our analysis reveals a clear distinction between safe and unsafe prompts in the intermediate layer representation space of the LLM, demonstrating the model's inherent ability to differentiate between these categories. However, a surprising and critical finding is the positioning of jailbreak attack prompts. These prompts are located near the boundary between safe and unsafe clusters, occupying a precarious position that allows them to potentially confuse the LLM. This strategic placement enables jailbreak attacks to bypass the model's built-in safety mechanisms, potentially leading to the generation of harmful content. The boundary position of jailbreak attack prompts presents a significant vulnerability in LLMs. Due to their ambiguous location, these prompts may be mistakenly classified within the safe space. This misclassification can induce the LLM to incorrectly judge harmful prompts as harmless and respond to them accordingly, thereby compromising the model's safety guardrails and producing potentially dangerous outputs.
LLMGuardrail effectively shifts the representations of jailbreak attack prompts from their ambiguous boundary position back into the clearly defined unsafe space. This recalibration is visually represented by the yellow points in Figure \ref{fig:exp_jailbreak_tsne}, which show the adjusted positions of the jailbreak prompts after the application of LLMGuardrail. By repositioning these prompts, LLMGuardrail enables the model to correctly identify them as potentially harmful, preventing misjudgment and inappropriate responses.

The ability of LLMGuardrail to recalibrate prompt representations in the model's embedding space is crucial for mitigating harmful behavior and improving the overall quality and safety of generated content. This recalibration allows the model to properly defend against or refuse to answer unsafe queries, thereby significantly enhancing its safety and reliability. Furthermore, this shift demonstrates LLMGuardrail's capacity to adjust the model's internal representations, making it more robust against inputs specifically designed to bypass safety measures.

\section{Conclusion}
In this paper, we presented LLMGuardrail, a novel framework for obtaining unbiased steering representations in large language models (LLMs) by incorporating causal analysis and adversarial learning techniques. Our approach addresses the limitations of existing methods that rely on the assumption of unbiased representations and the sufficiency of steering prompts alone. By systematically identifying and blocking the confounding effects of semantic biases, LLMGuardrail enables the extraction of steering representations that accurately capture the desired attributes or concepts.

\bibliographystyle{ACM-Reference-Format}
\balance
\bibliography{llmguardaril}


\end{document}